\documentclass[letterpaper]{article} 
\usepackage{aaai25}  
\usepackage{times}  
\usepackage{helvet}  
\usepackage{courier}  
\usepackage[hyphens]{url}  
\usepackage{graphicx} 
\usepackage{natbib}  
\usepackage{caption} 
\usepackage{amsmath}
\usepackage{amsthm}
\usepackage{amssymb}
\usepackage{booktabs}
\usepackage{multirow}

\usepackage{subfig}

\usepackage{colortbl}
\usepackage{algorithmic}
\usepackage[ruled,linesnumbered,vlined]{algorithm2e}

\usepackage{newfloat}
\usepackage{listings}
\DeclareCaptionStyle{ruled}{labelfont=normalfont,labelsep=colon,strut=off} 
\title{Out-of-Distribution Generalization on Graphs via Progressive Inference}

\author{
    Yiming Xu\textsuperscript{\rm 1,2},
    Bin Shi\textsuperscript{\rm 1,2}\thanks{Corresponding author.},
    Zhen Peng\textsuperscript{\rm 1,2},
    Huixiang Liu\textsuperscript{\rm 1,2},
    Bo Dong\textsuperscript{\rm 2,3},
    Chen Chen\textsuperscript{\rm 4}
}
\affiliations{
    \textsuperscript{\rm 1}School of Computer Science and Technology, Xi'an Jiaotong University\\
    \textsuperscript{\rm 2}Shaanxi Provincial Key Laboratory of Big Data Knowledge Engineering, Xi’an Jiaotong University\\
    \textsuperscript{\rm 3}School of Distance Education, Xi’an Jiaotong University\\
    \textsuperscript{\rm 4}University of Virginia, Charlottesville, Virginia, USA\\

    \{xym0924, liuhxwork\}@stu.xjtu.edu.cn, \{shibin, dong.bo\}@xjtu.edu.cn, zhenpeng27@outlook.com, zrh6du@virginia.edu
}

\usepackage{bibentry}

\begin{document}

\maketitle

\begin{abstract}
The development and evaluation of graph neural networks (GNNs) generally follow the independent and identically distributed (i.i.d.) assumption. Yet this assumption is often untenable in practice due to the uncontrollable data generation mechanism. In particular, when the data distribution shows a significant shift, most GNNs would fail to produce reliable predictions and may even make decisions randomly. One of the most promising solutions to improve the model generalization is to pick out causal invariant parts in the input graph. Nonetheless, we observe a significant distribution gap between the causal parts learned by existing methods and the ground-truth, leading to undesirable performance. In response to the above issues, this paper presents GPro, a model that learns graph causal invariance with progressive inference. Specifically, the complicated graph causal invariant learning is decomposed into multiple intermediate inference steps from easy to hard, and the perception of GPro is continuously strengthened through a progressive inference process to extract causal features that are stable to distribution shifts. We also enlarge the training distribution by creating counterfactual samples to enhance the capability of the GPro in capturing the causal invariant parts. Extensive experiments demonstrate that our proposed GPro outperforms the state-of-the-art methods by 4.91\% on average. For datasets with more severe distribution shifts, the performance improvement can be up to 6.86\%. \footnote{The code and data are available at: https://github.com/yimingxu24/GPro.}
\end{abstract}

%

\section{Introduction}
The powerful graph representation learning abilities of graph neural networks (GNNs) have been widely acknowledged in both academia and industry, and have been proven to be effective in a variety of applications, such as recommender systems~\cite{niu2020dual,xia2022multi,seo2022siren,yan2023cascading}, finance~\cite{liu2021pick,zhang2022efraudcom,shi2023edge,zheng2023survey}, life sciences~\cite{hsieh2021drug,zhu2022neural,su2022biomedical,fu2023spatial} and autonomous driving~\cite{gao2020vectornet,xu2022adaptive}. Despite their remarkable success, existing GNNs typically rely on the assumption that training and testing data are independently and identically distributed (i.i.d.). However, this assumption often becomes untenable in realistic scenarios due to the uncontrollable underlying data generation mechanism~\cite{bengio2019meta, li2022out}. 
Several recent studies have revealed the vulnerability of GNNs in the face of differently distributed data~\cite{ding2021closer, gui2022good}. The lack of out-of-distribution (OOD) generalization capabilities hinders the deployment of GNNs in multiple high-risk scenarios in the open world.

Recently, one of the most promising directions for improving out-of-distribution (OOD) generalization is the method based on causal invariant learning. 
Specifically, most existing studies~\cite{sui2022causal,fan2022debiasing,wu2022discovering} obtain node representations by GNNs and identify causal invariant substructures and features from the input graph in a single-step manner, such as directly applying dot product operations or MLPs. Finally, they introduce specialized optimization objectives and constraints to minimize the risk of causal invariance across different distributions. However, the attention in existing works has focused on the design of optimization objectives, but ignored the exploration of model architectures. Unlike grid-like data, the intricate nature of graphs presents a substantial challenge to this problem since the topological structure leads to complex coupling associations between the causal and non-causal parts. This challenge raises a serious concern: \textit{How powerful is this kind of single-step manner in uncovering causal substructures in the OOD scenarios?}

\begin{figure}
  \centering
  \includegraphics[width=1\linewidth]{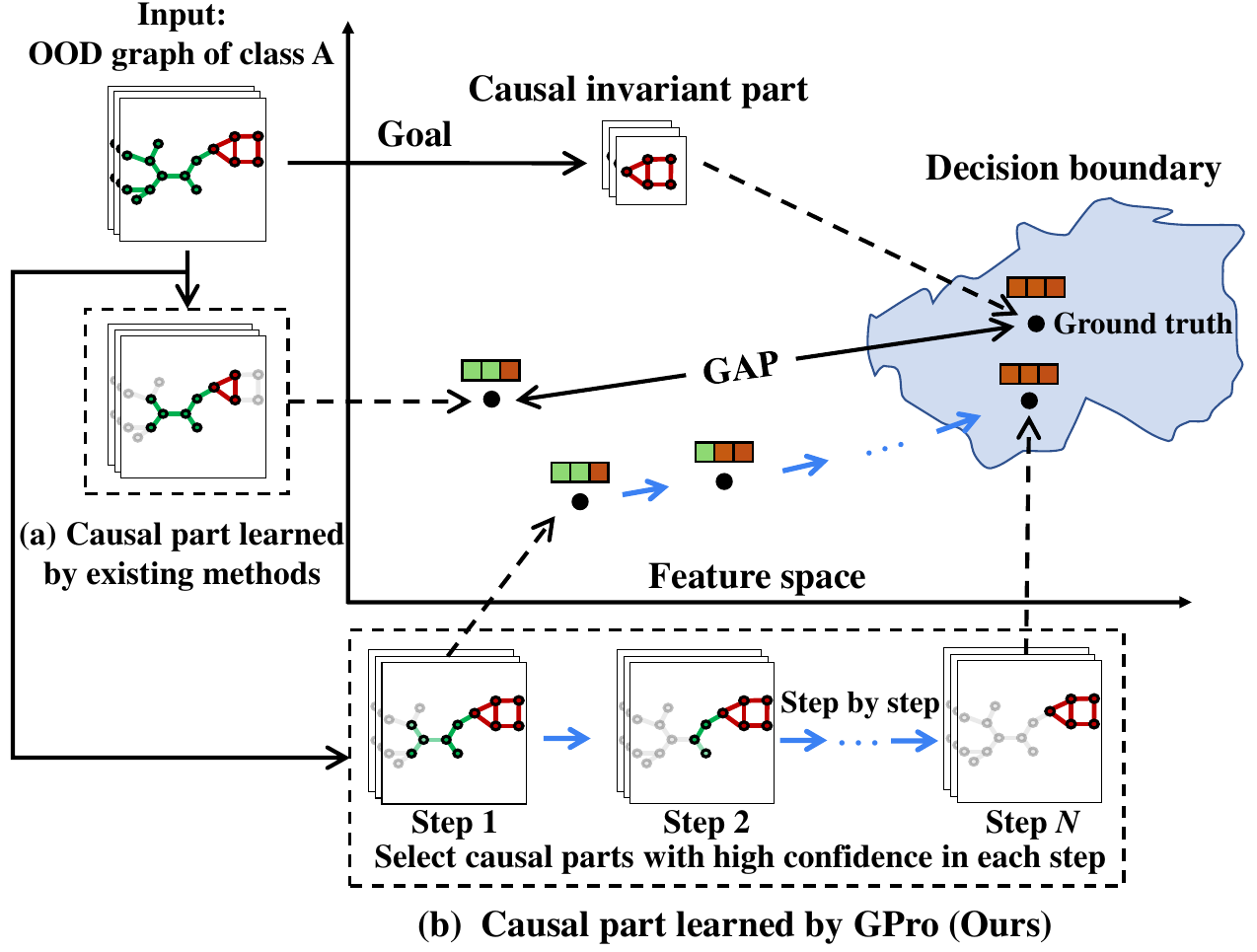}
  \caption{An illustration of the differences between existing methods and our proposed solution GPro. (a) The standard methods incorporate a significant amount of non-causal information (the green part in the input) in the learned features, resulting in a deviation from the decision boundary. (b) Our method is continuously refined via progressive inference to approach the ground truth.}
  \label{fig:toy_example}
\end{figure}

To validate this concern, we conduct an empirical study to investigate the effectiveness of existing methods in tackling this challenge. Specifically, in an OOD dataset, we visualize the causal features learned by existing methods and the ground-truth causal features (learned only by feeding the causal substructures into the GNN model) in the feature space. Unfortunately, our findings reveal a significant distribution gap between these two sets of features (details in Figure~\ref{fig:gap} and Figure~\ref{fig:tsne}). In other words, existing methods fail to capture high-quality causal invariant features, adversely affecting the generalization ability of the model. Indeed, when tackling complex problems, humans typically rely on multi-step inference rather than expecting immediate accurate results. For example, mathematicians break down difficult proofs into a series of sub-proofs and iteratively advance from intermediate results to achieve conclusive solutions. Inspired by this insight, we explore the encoder architecture based on the progressive inference paradigm on graphs to emulate the cognitive processes employed by humans when solving complex problems, aiming to enhance generalization capabilities, as illustrated in Figure~\ref{fig:toy_example}. 

In this paper, we present a new framework to learn \textbf{G}raph causal invariance via \textbf{Pro}gressive inference, called GPro, which decomposes the complex problem of discerning causal substructures and features into multiple intermediate inference steps from easy to hard. Specifically, each inference step further separates out the non-causal substructure with high confidence from the intermediate result learned in the previous inference step via an attention-based substructure context inference block. By stacking multiple such blocks, GPro mimics a step-by-step thought process to refine an accurate answer. Since the causal and non-causal parts are complementary, instead of only focusing on identifying the causal substructures, GPro employs a dual-tower model that concurrently identifies the causal and non-causal substructures, which aims to facilitate mutual assistance. Furthermore, to make the progressive inference process better capture the causal invariant parts, we propose to enlarge the training distribution by constructing different counterfactual samples through two feature-level data augmentation techniques. We also propose a novel supervised contrastive learning loss in graph causal invariant learning that leverages the supervised signals within and between samples of a batch. Our main contributions are summarized as follows:

$\bullet$ We propose the new concept of progressive inference in graph out-of-distribution generalization, which transforms the invariant learning process into multiple inference steps. This overcomes the existing model's inability to effectively disentangle the complex coupling associations between causal and non-causal substructures limitations.

$\bullet$ We introduce sophisticated feature augmentation strategies to enlarge the training distribution by generating counterfactual samples. Moreover, we present a novel supervised contrastive learning objective that effectively utilizes inter-sample supervised signals to further enhance the generalization ability of the model.

$\bullet$ The experimental results demonstrate that our GPro produces state-of-the-art results on 11 established baselines, and outperforms the sub-optimal baseline by 4.91\% on average. Qualitative and quantitative analysis of progressive inference and ablation studies corroborate the effectiveness of each component in GPro.

\section{Related Work}
\label{sec:related}
Graph neural networks have demonstrated impressive performance in a variety of applications~\cite{qiu2018deepinf,wang2019kgat,fu2022federated,wang2022molecular,xue2022quantifying,xu2023cldg,xu2024learning,fu2024federated,xu2025ted}. However, most existing methods fail in terms of model generalization, which hinders the deployment of GNNs in high-risk applications in the open world.
Recent studies are exploring how to improve the generalizability of GNNs in OOD scenarios, with efforts focusing on data-centric methods and causal invariant learning approaches. 
Data-centric methods~\cite{sui2022adversarial,li2023graph} improve OOD generalization ability through data augmentation.
Causal invariant learning methods~\cite{wu2022discovering,li2022learning,li2022ood} emphasize minimizing causal invariant risks in different distributions by introducing specialized optimization objectives and constraints.
For example, StableGNN~\cite{fan2023generalizing} extracts causal structures from input graphs to help the model eliminate spurious correlations. 
CAL~\cite{sui2022causal} and DisC~\cite{fan2022debiasing} divide the input graph into causal and non-causal graphs, and encourages a stable relationship between causal estimates and predictions.
CIGA~\cite{chen2022learning} proposes an information-theoretic objective to capture the invariance of graphs to guarantee OOD generalization under various distributional shifts.
FLOOD~\cite{liu2023flood} constructs multiple environments from graph data augmentation and learns invariant representation under risk extrapolation. For more extensive work, please refer to~\cite{li2022out}.
Although these methods show higher effectiveness, they still suffer from at least one of the following limitations: 
(1) Prior works ignore the important role of encoder architectures in OOD generalization. \cite{chen2022learning} highlights that it is promising to obtain better OOD generalization ability by incorporating more advanced architectures. As shown in Figure~\ref{fig:gap} and Figure~\ref{fig:tsne}, we confirm that existing methods are not sufficient to deal with this complex problem. 
(2) Some methods ignore the important role of increasing the diversity of training data, i.e., enlarging the training distribution, to improve generalization performance. (3) Existing methods do not fully consider supervised signals that exist within and between samples in a batch. Overall, the above limitations lead to sub-optimal solutions.

\begin{figure*}
  \centering
  \includegraphics[width=1\linewidth]{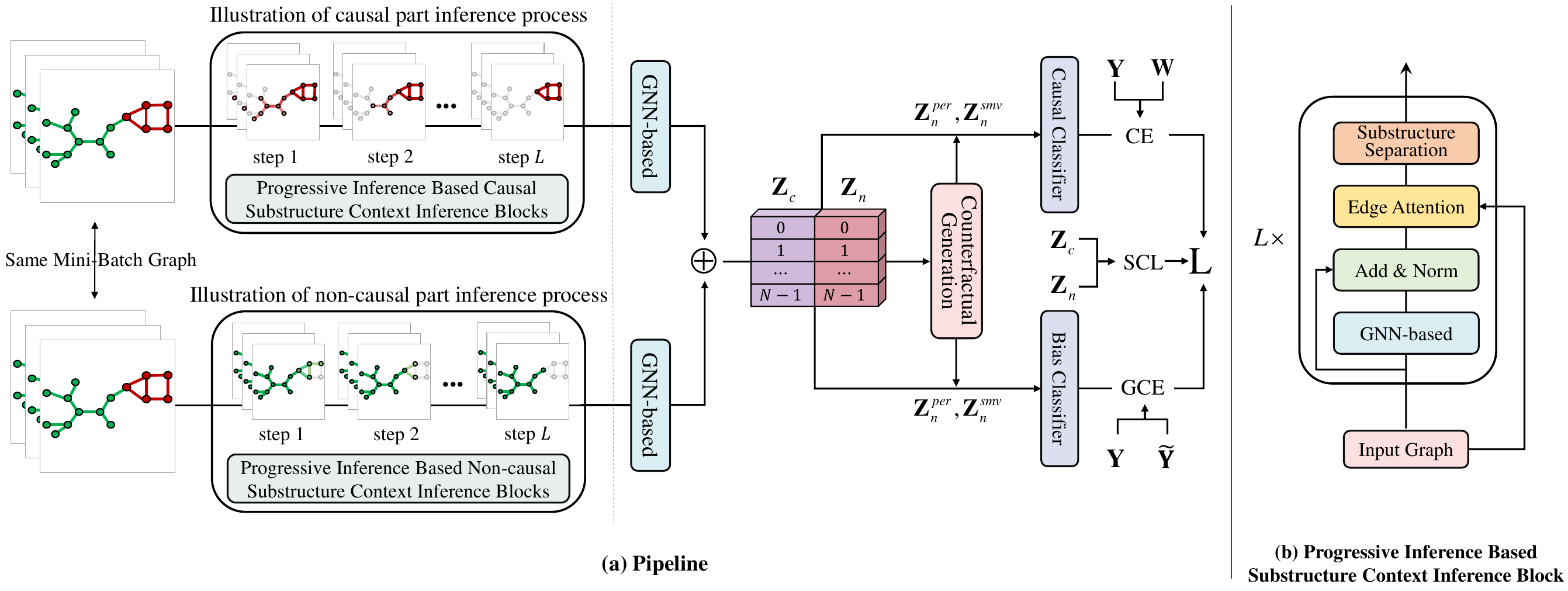}
  \caption{The pipeline and implementation details of the GPro. The basic idea is to decompose the complex problem of causal invariant learning on graphs into multiple intermediate inference steps, and finally extract causal features with generalization through progressive inference. Notably, in the input graph toy example of the leftmost, the red part and the green part are defined as causal and non-causal substructures.}
  \label{fig:overview}
\end{figure*}

\section{Methodology}
The pipeline of GPro is shown in Figure~\ref{fig:overview}. It consists of three major components: progressive inference-based substructure context inference block, counterfactual graph sample generation, and causal learning loss function. First, the substructure context inference block extracts causal and non-causal representations via step-by-step inference. Then, two strategies are designed to generate counterfactual samples to enlarge the training distribution. Finally, the loss function promotes a causal relationship between causal representations and labels, while eliminating any misleading correlations between non-causal representations and labels.

\subsection{Problem Formulation}
Suppose we are given training and testing graph data $\mathcal{G}_{train}=\left\{ \left ( G_{i},Y_{i} \right )\right\}_{i=1}^{N^{tr}}$ and $\mathcal{G}_{test}=\left\{ \left ( G_{i},Y_{i} \right )\right\}_{i=1}^{N^{te}} $, drawn from distributions $P\left ( \mathcal{G}_{train} \right )$ and $P\left ( \mathcal{G}_{test} \right )$, respectively. $\mathcal{G}_{test}$ is unobserved in the training stage. In the out-of-distribution setting, our goal is to learn a graph predictor $f$ that achieves a satisfactory generalization on a testing set with an unknown distribution:

\begin{equation}
f_{\theta }=\arg{\underset{f_{\theta }}{\min{}}}\mathbb{E}_{G,Y \sim P\left ( \mathcal{G}_{test} \right )}\left [ \ell \left ( f_{\theta }\left ( G \right ),Y \right ) \right ],
\end{equation}
where the distribution shift exists in the training set and the unseen testing set, i.e., $P\left ( \mathcal{G}_{train} \right ) \neq  P\left ( \mathcal{G}_{test} \right )$, and $\ell \left ( \cdot, \cdot  \right ) : \mathbb{Y} \times \mathbb{Y} \to \mathbb{R}$ denotes a loss function.

\subsection{Substructure Context Inference Block}
To address the challenges posed by the complex topology of graphs for causal invariant learning, we decompose the complex inference problem of learning causal structures and features into multiple intermediate inference steps. 
Since the causal and non-causal parts are complementary, we employ a dual-tower model where one tower is responsible for identifying the non-causal part, while the other tower focuses on recognizing the causal part. These two towers work in tandem and provide mutual assistance to each other in the overall task.
The illustration of our proposed GPro and its implementation details are shown in Figure~\ref{fig:overview}. 

Specifically, given an input graph $G=\left\{ \mathbf{A}, \mathbf{X} \right\}$, where $\mathbf{A}$ and $\mathbf{X}$ are the adjacency matrix and node features, respectively, we first employ an edge attention layer to measure the causal importance of edges, and edge-level attention scores are estimated by considering three simple but effective encodings, namely GNN update node feature encoding, node centrality encoding, and inter-node similarity encoding. The node features are updated by a GNN encoder, and employ a residual connection~\cite{he2016deep} and batch normalization~\cite{ioffe2015batch} following the GNN layer: 
\begin{equation}
\label{eq:gnn}
\mathbf{H} = f\left ( \mathbf{A},\mathbf{X} \right ). 
\end{equation}

Then, unlike previous methods that ignore node centrality, we realize the role of node centrality in measuring the importance of nodes~\cite{ying2021transformers}, and additionally introduce the degree centrality of nodes to comprehensively portray their representations.
\begin{equation}
\mathbf{q}_i = \textrm{MLP}_{\textrm{node}}\left ( \left [ \left| \mathcal{N}\left ( i \right )\right| ; \mathbf{h}_i \right ] \right ),
\end{equation}
where $\left| \mathcal{N}\left ( i \right ) \right|$ is the degree of node $i$, $ \mathbf{h}_i=\mathbf{H}\left [ \textrm{i},: \right ]$ is the feature of node $i$ updated by the GNN encoder $f$, and $\left [ ; \right ]$ is the concatenation operation. The inter-node similarity is encoded through $\textrm{sim} \left ( \cdot, \cdot  \right )$. Finally, the calculation formula of edge-level attention $\alpha_{ij}$ for node $i$ and node $j$ is as follows:
\begin{equation}
\label{eq:attention-score}
\alpha_{ij} = \sigma \left ( \textrm{MLP}_{\textrm{edge}}([ \textrm{sim}\left ( \mathbf{q}_i,\mathbf{q}_j \right ); \mathbf{q}_i; \mathbf{q}_j]) \right ),
\end{equation}
where $\alpha_{ij}\in \left ( 0,1 \right ) $ denotes the edge-level attention score of edge~$\left ( i,j \right )$ in the causal substructure. $\sigma\left ( \cdot  \right )$ is the sigmoid function. Additionally, we define $\textrm{sim}\left ( \mathbf{q},\mathbf{k} \right )=\mathbf{q}^{T}\mathbf{k}/\left\|\mathbf{q} \right\|\left\|\mathbf{k} \right\|$.

To separate causal substructures and features from the original graph step-by-step, at each inference step, the substructure separation layer constructs a mask matrix $\mathbf{M}$ to further separate the $\rho$ (e.g., 10\%) substructures with the lowest score in the causal attention score matrix $\mathbf{E}$, where $\mathbf{E}\left [ \textrm{i},\textrm{j} \right ]=\alpha_{ij}$, i.e., those should belong to the non-causal part, from the causal substructures learned from the previous inference step:
\begin{equation}
\label{eq:mask}
\mathbf{M} = \textrm{rank}\left ( \mathbf{E}, \left \lfloor \rho\left|\mathcal{E} \right|  \right \rfloor \right ),
\end{equation}
where $\mathbf{M}=\left\{ 0,1\right\}^{\left| \mathcal{V}\right| \times \left| \mathcal{V}\right|}$, $\left| \mathcal{V}\right|$ and $\left| \mathcal{E}\right|$ are the number of nodes and edges in the graph $G$, respectively. The $\textrm{rank}$ function sorts the causal attention scores in $\mathbf{E}$, $\mathbf{M}_{ij}=1/0$ indicates that the edge~$\left ( i,j \right )$ is determined to be the causal/non-causal substructure in the current inference step.

Afterward, we update the adjacency matrix and edge-level attention scores in the next intermediate inference step through the mask matrix $\mathbf{M}$ constructed by the $ \left ( l-1 \right )$-th layer intermediate inference step:
\begin{equation}
\label{eq:sep}
\mathbf{A}^{l} = \mathbf{A}^{l-1}\odot \mathbf{M},
\end{equation}
where $\odot$ denotes the Hadamard product of the matrix.

In the progressive inference process, each intermediate inference step is modeled by a substructure context inference block, involving Eq.~\eqref{eq:gnn} to Eq.~\eqref{eq:sep}, as illustrated in Figure~\ref{fig:overview}(b). We obtain more reliable causal substructure $G_{c}=\left\{ \mathbf{A}_{c}^{L},\mathbf{H}_{c}^{L}\right\}$ and non-causal substructure $G_{n}=\left\{ \mathbf{A}_{n}^{L},\mathbf{H}_{n}^{L}\right\} $ through an $L$-step intermediate inference process, that is, stacking $L$ layers of causal and non-causal substructure context inference blocks that do not share parameters. 
After deriving the final causal and non-causal substructures, we learn causal and non-causal graph-level representations through GNN encoders and the pooling operation:
\begin{align}
\label{eq:gcn1}
\mathbf{Z}_{c} &= f_{readout}\left ( f_{c}\left ( \mathbf{A}_{c}^{L},\mathbf{H}_{c}^{L} \right ) \right ), \\
\label{eq:gcn2}
\mathbf{Z}_{n} &= f_{readout}\left ( f_{n}\left ( \mathbf{A}_{n}^{L},\mathbf{H}_{n}^{L} \right ) \right ), 
\end{align}
where $f_{readout}\left ( \cdot  \right )$ is a readout function to generate the graph-level representation. $\mathbf{Z}_{c},\mathbf{Z}_{n}\in \mathbb{R}^{N\times d}$ are causal and non-causal representation matrices in the mini-batch graph, respectively. The batch size is $N$.

\subsection{Counterfactual Graph Sample Generation}
To this point, we have extracted causal and non-causal representations in graphs through a complex multi-step inference process. To further improve the graph OOD generalization, we employ two strategies to generate counterfactual graph representations to eliminate correlations between causal and non-causal variables, while increasing the diversity of samples and enlarging the training distribution.
As causal variables reflect invariant intrinsic properties in graph data, inappropriate interventions on causal representations may lead to changes in the semantics and labels of the input graph. However, there is no causality between non-causal representations and labels. Therefore, we could enlarge the training distribution through robust interventions on the non-causal representation. \looseness=-1

The first counterfactual graph representation generation strategy is to randomly permute the non-causal representations. Random permute has proven to be effective in OOD problems in several domains~\cite{lee2021learning,sui2022causal}. The $\textrm{permute}\left ( \cdot  \right ) $ function randomly permutes the order of the graphs in the mini-batch.
\begin{equation}
\label{eq:permute}
\textrm{idx} = \textrm{permute} \left ( N  \right ),
\end{equation}
where $\textrm{idx}$ is the new indices after random permutation. $\mathbf{Z}_{n}^{per}$ is the randomly permute non-causal representation matrix, i.e., $\mathbf{Z}_{n}^{per}=\mathbf{Z}_{n}\left [ \textrm{idx},: \right ]$.

Inspired by~\cite{tang2021crossnorm}, we design a new counterfactual sample generation strategy for graph-level representations. The core of the second strategy is to enlarge the training distribution by swapping the mean and variance between the non-causal representations of the samples in the mini-batch. 
\begin{equation}
\label{eq:swap}
\mathbf{Z}_{n}^{smv}=\sigma _{\mathbf{Z}_{n}^{per}}\frac{\mathbf{Z}_{n}-\mu_{\mathbf{Z}_{n}} }{\sigma_{\mathbf{Z}_{n}} } + \mu _{\mathbf{Z}_{n}^{per}},
\end{equation}
where $\mu_{\mathbf{Z}_{n}},\sigma_{\mathbf{Z}_{n}} $ are the means and variances of the non-causal representations of each sample in the minibatch, and $\mu_{\mathbf{Z}_{n}^{per}},\sigma_{\mathbf{Z}_{n}^{per}} $ are the means and variances of the non-causal graph representations after random permutation.

\subsection{Causal Learning Loss Function}
It is necessary to design reasonable loss functions to ensure causal relationships between causal features and labels while eliminating spurious correlations between non-causal features and labels. After counterfactual graph sample generation, given a mini-batch of graphs, we can extract three graph-level representations, i.e., a real graph representation $\mathbf{Z}=\left [ \mathbf{Z}_{c};\mathbf{Z}_{n} \right ]$ and two counterfactual graph representations $\mathbf{Z}^{per}=\left [ \mathbf{Z}_{c};\mathbf{Z}_{n}^{per} \right ]$, and $\mathbf{Z}^{smv}=\left [ \mathbf{Z}_{c};\mathbf{Z}_{n}^{smv} \right ]$. 
Since the causal and non-causal parts are complementary, we employ a dual-tower model to identify the causal and non-causal parts, respectively. Therefore, we firstly design two classifiers, namely causal classifier $\Phi_{c}$ and non-causal classifier $\Phi_{n}$ to train this dual-tower model (note that, the loss from $\Phi_{c}$ is not back-propagated to the encoder model involved in generating non-causal features, and vice versa).
The purpose of the causal branch is to estimate causal features, so we classify its representation to the ground-truth label. Thus, we define the supervised classification loss as cross-entropy (CE) loss to train the causal encoder. Meanwhile, we utilize the generalized cross-entropy (GCE)~\cite{zhang2018generalized} loss and target labels to train a non-causal encoder and classifier. GCE loss is described as:
\begin{equation}
\textrm{GCE}\left ( \Phi_{n}\left ( \mathbf{z} \right ),\mathbf{y} \right )=\frac{1-\Phi_{n}^{y}\left ( \mathbf{z} \right )^{q}}{q},
\end{equation}
where $y$ refers to the ground truth label, $\Phi_{n}\left ( \mathbf{z} \right )$ and $\Phi_{n}^{y}\left ( \mathbf{z} \right )$ indicate the softmax output of the non-classifier $ \Phi_{n}$ and its probability belonging to the target class $y$, respectively. $q$ is a hyperparameter. The GCE loss imposes a higher weight on the gradient of the CE loss for samples, which have high confidence $\Phi_{n}^{y}$ of the target category $y$. It is defined as follows: 
\looseness=-1

\begin{equation}
\frac{\partial GCE(\Phi_{n}\left ( \mathbf{z} \right ),y)}{\partial\theta_n}={(\Phi_{n}^{y})^q}\frac{(\Phi_{n}\left ( \mathbf{z} \right ),y)}{\partial\theta_n},
\end{equation}
where non-causal shortcut information is usually easier to learn and will have larger ${(\Phi_{n}^{y})^q}$ as confirmed by prior work~\cite{lee2021learning,fan2022debiasing}. GCE loss amplifies the gradient by ${(\Phi_{n}^{y})^q}$ to emphasize the non-causal encoder and classifier $\Phi_{n}$ overfocus on non-causal information. Therefore, we train the causal and non-causal parts with CE and GCE losses, respectively.
The mathematical definition of the objective function is as follows:
\begin{equation}
\label{eq:loss_1}
\mathcal{L}_{\textrm{dis}} = \textrm{CE}\left ( \Phi_{c}\left ( \mathbf{Z} \right ),\mathbf{Y} \right ) + \textrm{GCE}\left ( \Phi_{n}\left ( \mathbf{Z} \right ),\mathbf{Y} \right ).
\end{equation}

In addition, we also train the causal and non-causal encoders by the CE and GCE loss between the counterfactual graph representations $\mathbf{Z}^{per}$, $\mathbf{Z}^{smv}$ and the target labels, respectively. 
For the causal part, we maintain the consistency between causal features and the target label $\mathbf{Y}$, which is equivalent to expanding the training distribution, thereby better training the causal classifier.
To make the spurious correlation between counterfactual graph representations and labels still exist, we permute the label $\widetilde{\mathbf{Y}} = \mathbf{Y}\left [ \textrm{idx} \right ]$ along with $\mathbf{Z}^{per}$ and $\mathbf{Z}^{smv}$ as the target labels for the output of $\Phi_{n}$.
This ensures that the non-causal encoder and classifier continuously focus on the non-causal information.
Meanwhile, samples can be regarded as unbiased and high quality when the loss of the causal classifier is small, but the loss of the non-causal classifier is large. Inspired by~\cite{lee2021learning}, we enforce the causal encoder and classifier to learn causality by increasing the weights of counterfactual samples of unbiased samples by $ \mathbf{W}\left ( \mathbf{Z} \right ) = \frac{\textrm{CE}\left ( \Phi_{n}\left ( \mathbf{Z} \right ),\mathbf{Y} \right )}{\textrm{CE}\left ( \Phi_{c}\left ( \mathbf{Z} \right ),\mathbf{Y} \right ))+\textrm{CE}\left ( \Phi_{n}\left ( \mathbf{Z} \right ),\mathbf{Y} \right )}$. Moreover, $\mathcal{L}_{\textrm{cou}}$ is not used during the initial training phase because the generated counterfactual graph representations are of low quality and may lead to label changes. $\mathcal{L}_{\textrm{cou}}$ is formally defined as follows:
\begin{equation}
\label{eq:loss_2}
\resizebox{1\hsize}{!}{$
\begin{aligned}
\mathcal{L}_{\textrm{cou}} &= \mathbf{W}\left ( \mathbf{Z} \right )\left ( \textrm{CE}\left ( \Phi_{c}\left ( \mathbf{Z}^{per} \right ),\mathbf{Y} \right )+ \textrm{CE}\left ( \Phi_{c}\left ( \mathbf{Z}^{smv} \right ),\mathbf{Y} \right )\right )/2 \\
&+ \left ( \textrm{GCE}\left ( \Phi_{n}\left ( \mathbf{Z}^{per} \right ),\widetilde{\mathbf{Y}} \right ) + \textrm{GCE}\left ( \Phi_{n}\left ( \mathbf{Z}^{smv} \right ),\widetilde{\mathbf{Y}} \right )\right )/2.
\end{aligned}
$}
\end{equation}

To enhance the disentanglement between causal and non-causal representations, a novel loss function is proposed in this work, which extends supervised contrastive learning~\cite{khosla2020supervised} (SCL) into the graph causal invariant learning. Specifically, by leveraging the label information, the proposed method pulls together causal graph representations that belong to the same class in a batch, while pushing apart causal graph representations from different classes and non-causal graph representations from all classes. The novel supervised contrastive loss of graph causal invariance principle is defined as follows:
\begin{equation}
\label{eq:loss_3}
\resizebox{1\hsize}{!}{$
\mathcal{L}_{\textrm{scl}}=\sum_{i\in I}\frac{-1}{\left|P\left ( i \right ) \right|} \log{\frac{\sum\limits_{p\in P\left ( i \right )}\exp{\left ( \mathbf{z}_{i}^{c}\cdot \mathbf{z}_{p}^{c}/\tau  \right )}}{\sum\limits_{j\in A\left ( i \right )}\exp{\left ( \mathbf{z}_{i}^{c}\cdot \mathbf{z}_{j}^{c}/\tau  \right )}+{\sum\limits_{k\in I}\exp{\left ( \mathbf{z}_{i}^{c}\cdot \mathbf{z}_{k}^{n}/\tau  \right )}}}},
$}
\end{equation}
where $i\in I\equiv \left\{ 1...N\right\}$ is the index in the mini-batch, and $A\left ( i \right ) \equiv I \setminus \left\{ i\right\} $. $P\left ( i \right )\equiv \left\{ p\in A\left ( i \right ):\boldsymbol{y}_p = \boldsymbol{y}_i\right\}$ is the set of indices that have the same label as graph $i$, and $\tau$ is a temperature parameter. $\mathbf{z}_{i}^{c}$ and $\mathbf{z}_{k}^{n}$ are the causal and the non-causal representation of graph $i$ and $k$, respectively. 

Note that the causal and non-causal substructure context inference blocks have the same architecture but do not share weights, we expect both encoders to make similar judgments on edge-level attention scores. We impose a consistency constraint on the context inference blocks of causal and non-causal substructures via mean squared error (MSE) loss. \looseness=-1

\begin{equation}
\label{eq:loss_4}
\mathcal{L}_{\textrm{con}}=\textrm{MSE}\left ( \mathbf{E}_{c},\mathbf{E}_{n} \right ),
\end{equation}
where $\mathbf{E}_{c}$ and $\mathbf{E}_{n}$ are the learned attention score matrices for the causal and non-causal substructure context downsampling blocks, respectively. 

Finally, combining all the above defined losss functions, the total causal learning loss function is defined as:
\begin{equation}
\label{eq:loss_function}
\mathcal{L}=\mathcal{L}_{\textrm{dis}}+\lambda_{1}\mathcal{L}_{\textrm{cou}}+\lambda_{2}\mathcal{L}_{\textrm{scl}}+\lambda_{3}\mathcal{L}_{\textrm{con}},
\end{equation}
where $\lambda_{1}$, $\lambda_{2}$, and $\lambda_{3}$ are hyperparameters for weighing the importance of counterfactual loss, supervised contrastive loss, and consistency loss, respectively. 
The details of our algorithm are summarized in the Appendix.

\begin{table*}[!htbp]
\renewcommand\arraystretch{1.24}
\resizebox{1\textwidth}{!}{
 \centering
\begin{tabular}{c|ccc|ccc|ccc}
\hline \hline
\multicolumn{1}{c|}{Dataset}                                         & \multicolumn{3}{c|}{CMNIST-75sp}                                                                                                                        & \multicolumn{3}{c}{CFashion-75sp}     & \multicolumn{3}{|c}{CKuzushiji-75sp}                                                                                        \\ 
\multicolumn{1}{c|}{Bias}                                            & \multicolumn{1}{c}{0.8}               & \multicolumn{1}{c}{0.9}                & \multicolumn{1}{c|}{0.95} & \multicolumn{1}{c}{0.8}                             & \multicolumn{1}{c}{0.9}                 & \multicolumn{1}{c|}{0.95} & \multicolumn{1}{c}{0.8}                             & \multicolumn{1}{c}{0.9}                 & \multicolumn{1}{c}{0.95} \\ \hline
 GCN~\cite{kipf2016semi}                & $50.43_{\pm 4.13}$ & $28.97_{\pm 4.40}$  & $13.50_{\pm 1.38}$        & $63.60_{\pm 0.53}$  & $57.22_{\pm 0.93}$  & $47.69_{\pm 0.42}$  &  $38.45_{\pm 1.1}$ & $28.35_{\pm 0.79}$ &$20.70_{\pm 0.88}$ \\
GIN~\cite{xu2018powerful}                & $57.75_{\pm 0.78}$ & $36.78_{\pm 5.55}$ & $16.04_{\pm 1.14}$        & $64.25_{\pm 0.46}$ & $58.03_{\pm 0.40}$  & $49.74_{\pm 0.60}$ &  $41.83_{\pm 0.78}$ &   $30.09_{\pm 0.87}$ &   $21.18_{\pm 1.63}$    \\ 
 GCNII~\cite{chen2020simple}             & $69.70_{\pm 1.73}$   & $57.68_{\pm 1.68}$ & $41.00_{\pm 3.75}$         & $66.68_{\pm 0.59}$ & $60.58_{\pm 0.28}$  & $53.18_{\pm 0.08}$   & $48.53_{\pm 0.25}$ & $36.23_{\pm 0.20}$  & $25.60_{\pm 0.76}$ \\
FactorGCN~\cite{yang2020factorizable}                                            & $72.30_{\pm 1.18}$                     & $62.35_{\pm 5.07}$                 &    $42.50_{\pm 4.91}$                        & $61.23_{\pm 1.11}$  &$53.50_{\pm 1.29}$ &$45.78_{\pm 2.40}$    &  $42.87_{\pm 1.19}$   &      $32.35_{\pm 2.79}$ & $23.87_{\pm 0.12}$                \\
DiffPool~\cite{ying2018hierarchical}                                             & $73.79_{\pm 0.02}$                      & $66.45_{\pm 0.78}$                  &     $47.12_{\pm 1.04}$                       & $62.82_{\pm 0.53}$   &$57.50_{\pm 0.39}$ &$50.86_{\pm 0.20}$     &     $45.46_{\pm 0.65}$   &      $36.18_{\pm 0.19}$  & $27.45_{\pm 0.26}$                 \\ \hline
StableGNN~\cite{fan2023generalizing}                                            & $77.65_{\pm 1.64}$                      & $68.87_{\pm 1.74}$                   &     $51.33_{\pm 0.87}$                        & $64.03_{\pm 0.29}$   &$58.26_{\pm 0.09}$ &$51.46_{\pm 0.39}$     &      $49.41_{\pm 0.09}$   &      $39.30_{\pm 0.12}$   & $28.26_{\pm 0.14}$                  \\
$\text{LDD}_{GCN}$~\cite{lee2021learning}  & $64.95_{\pm 1.22}$   & $56.65_{\pm 2.18}$ & $46.83_{\pm 2.88}$        & $63.85_{\pm 1.17}$  & $64.30_{\pm 0.89}$   & $62.28_{\pm 0.48}$ & $42.38_{\pm 0.33}$ & $38.75_{\pm 0.49}$ & $33.08_{\pm 0.59}$ \\
$\text{LDD}_{GIN}$~\cite{lee2021learning}   &  $64.88_{\pm 1.45}$ & $50.59_{\pm 1.07}$ & $31.23_{\pm 2.48}$        & $64.65_{\pm 0.63}$ & $57.10_{\pm 0.43}$  & $53.38_{\pm 0.47}$ &   $37.83_{\pm 0.54}$ & $28.97_{\pm 0.18}$ & $22.13_{\pm 0.34}$ \\
$\text{LDD}_{GCNII}$~\cite{lee2021learning}   &  $78.03_{\pm 0.66}$ & $69.53_{\pm 0.96}$ & $51.05_{\pm 3.87}$        & $50.63_{\pm 1.79}$ & $54.09_{\pm 2.54}$ & $57.93_{\pm 0.88}$  & $48.70_{\pm 1.98}$  & $41.59_{\pm 1.07}$  & $33.93_{\pm 0.71}$ \\
$\text{CAL}_{GCN}$~\cite{sui2022causal}   & $77.10_{\pm 1.01}$   & $67.89_{\pm 0.45}$ & $51.42_{\pm 1.39}$        & $67.74_{\pm 0.31}$  & $60.90_{\pm 0.71}$  & $54.41_{\pm 0.15}$ & $52.18_{\pm 0.32}$ & $41.47_{\pm 0.69}$ &  $31.39_{\pm 0.65}$ \\ 
$\text{CAL}_{GIN}$~\cite{sui2022causal}     & $76.50_{\pm 0.40}$   & $65.32_{\pm 0.32}$  & $44.43_{\pm 1.28}$        & $65.04_{\pm 0.23}$ & $59.82_{\pm 0.39}$  & $52.98_{\pm 0.51}$ & $50.71_{\pm 0.41}$  & $38.40_{\pm 0.53}$ & $29.46_{\pm 0.49}$ \\ 
$\text{CAL}_{GAT}$~\cite{sui2022causal}     & $\underline{88.21_{\pm 0.50}}$ & $\underline{81.57_{\pm 0.21}}$ & $\underline{69.18_{\pm 1.10}}$        & $\underline{71.11_{\pm 0.06}}$  & $\underline{66.22_{\pm 0.36}}$ & $59.02_{\pm 0.39}$ & $\underline{64.54_{\pm 0.16}}$  & $\underline{52.00_{\pm 0.70}}$ & $\underline{37.93_{\pm 0.81}}$ \\ 
$\text{DisC}_{GCN}$~\cite{fan2022debiasing}   & $82.60_{\pm 0.93}$   & $78.14_{\pm 2.14}$ & $63.47_{\pm 5.65}$        & $66.85_{\pm 1.11}$  & $65.33_{\pm 4.70}$  & $\underline{63.93_{\pm 1.50}}$ & $55.53_{\pm 2.29}$ & $48.13_{\pm 2.59}$ &  $36.63_{\pm 1.73}$\\ 
$\text{DisC}_{GIN}$~\cite{fan2022debiasing}     & $82.10_{\pm 1.50}$   & $74.90_{\pm 1.81}$  & $58.58_{\pm 4.24}$        & $67.10_{\pm 1.07}$ & $59.90_{\pm 1.31}$  & $55.80_{\pm 0.36}$& $55.18_{\pm 1.00}$  & $41.75_{\pm 0.81}$& $30.25_{\pm 1.63}$ \\ 
$\text{DisC}_{GCNII}$~\cite{fan2022debiasing}     & $79.50_{\pm 2.48}$ & $76.00_{\pm 1.90}$ & $60.54_{\pm 5.33}$        & $66.47_{\pm 1.77}$  & $65.48_{\pm 0.70}$ & $61.75_{\pm 0.27}$ & $54.90_{\pm 1.30}$  & $44.73_{\pm 1.55}$& $36.95_{\pm 0.70}$ \\

$\text{CIGA}$~\cite{chen2022learning}     & $64.45_{\pm 3.49}$ & $48.56_{\pm 6.44}$ & $34.33_{\pm 2.63}$        & $59.37_{\pm 0.89}$  & $53.52_{\pm 1.98}$ & $45.37_{\pm 2.15}$ & $43.80_{\pm 2.46}$  & $31.74_{\pm 2.18}$& $22.89_{\pm 0.90}$ \\
$\text{GALA}$~\cite{chen2023does}     & $78.82_{\pm 1.66}$ & $64.73_{\pm 2.39}$ & $41.54_{\pm 3.25}$        & $65.64_{\pm 0.49}$  & $59.68_{\pm 1.47}$ & $51.72_{\pm 1.36}$ & $50.41_{\pm 1.70}$  & $33.69_{\pm 2.76}$& $24.16_{\pm 0.60}$ \\ \hline

$\text{GPro}$     & $\mathbf{88.87}_{\pm 1.03}$ & $\mathbf{87.58}_{\pm 0.36}$ & $\mathbf{79.34}_{\pm 1.07}$        & $\mathbf{75.41}_{\pm 0.36}$  & $\mathbf{70.57}_{\pm 0.29}$ & $\mathbf{64.72}_{\pm 0.71}$ & $\mathbf{66.46}_{\pm 0.56}$  & $\mathbf{58.35}_{\pm 0.63}$& $\mathbf{47.56}_{\pm 0.40}$ \\ 
\hline \hline
\end{tabular}}
\caption{Experimental results (\%) for the graph classification task on three datasets with unbiased testing sets. We report the mean accuracy and standard error. Bold indicates the optimal and underline indicates the suboptimal.}
\label{tab:main}
\end{table*}

\section{Experiments} \label{sec:sec5}
\subsection{Experiment Preparation}
\subsubsection{Datasets}
We use three benchmark graph classification datasets in causal learning~\cite{fan2022debiasing}, namely CMNIST-75sp, CFashion-75sp, and CKuzushiji-75s, to evaluate the performance of the models on out-of-distribution (OOD) problems. The datasets consider three bias degrees {0.8, 0.9, 0.95}, i.e., the causal and the non-causal substructures have 80\%, 90\%, and 95\% probabilities of co-occurrence in the training set. For example, at a bias degree of 0.9 in the training set of CMNIST-75sp superpixel graph, 90\% of the 0 digits come with a red background (i.e., biased samples), and the remaining 10\% come with a random background color (i.e., unbiased samples). Thus, it enables the establishment of spurious correlations between the non-causal substructures and the labels. The datasets are divided into the training set: validation set: testing set in the ratio of 10K:5K:10K. The testing sets are all unbiased samples. Each dataset contains 10 classes. Statistics of the datasets are provided in the Appendix.

\subsubsection{Baselines}
To verify that GPro produces consistent and significant improvements, we compare GPro with 11 state-of-the-art algorithms designed for in-distribution (ID) or out-of-distribution (OOD) learning. \textbf{In-Distribution Methods:} GCN~\cite{kipf2016semi}, GIN~\cite{xu2018powerful}, GCNII~\cite{chen2020simple}, FactorGCN~\cite{yang2020factorizable}, and DiffPool~\cite{ying2018hierarchical}.
\textbf{Out-of-Distribution Methods:} LDD~\cite{lee2021learning}, StableGNN~\cite{fan2023generalizing}, CAL~\cite{sui2022causal}, DisC~\cite{fan2022debiasing}, CIGA~\cite{chen2022learning} and GALA~\cite{chen2023does}. More details on the baselines can be found in the Appendix.

\subsubsection{Implementation Details} We use the Adam optimizer~\cite{kingma2014adam},
and the learning rate is 0.01. For Eq.~\eqref{eq:gcn1} and Eq.~\eqref{eq:gcn2}, we use the GCN~\cite{kipf2016semi} with 2 layers and 146 hidden dimensions as the encoder. We train the GPro with 200 epochs and add $\mathcal{L}_{\textrm{cou}}$ loss function at the 100th epoch. The batch size is 256. The default value for the number of causal and non-causal substructure context inference blocks is 2, and $\rho$ are 0.9 and 0.8, respectively. We set q of GCE loss as 0.7 to amplify the focus on the non-causal part, $\lambda_{1}$ is 15, $\lambda_{2}$ is 0.01 and $\lambda_{3}$ is 1. \looseness=-1

\subsection{Comparison with State-of-the-Art} \label{sec:sec5.2}
\begin{figure*}
    \setlength{\belowcaptionskip}{-0.4cm}
    \centering
    \subfloat[GCN]{\label{fig:gap1}\includegraphics[width=0.2\linewidth]{
    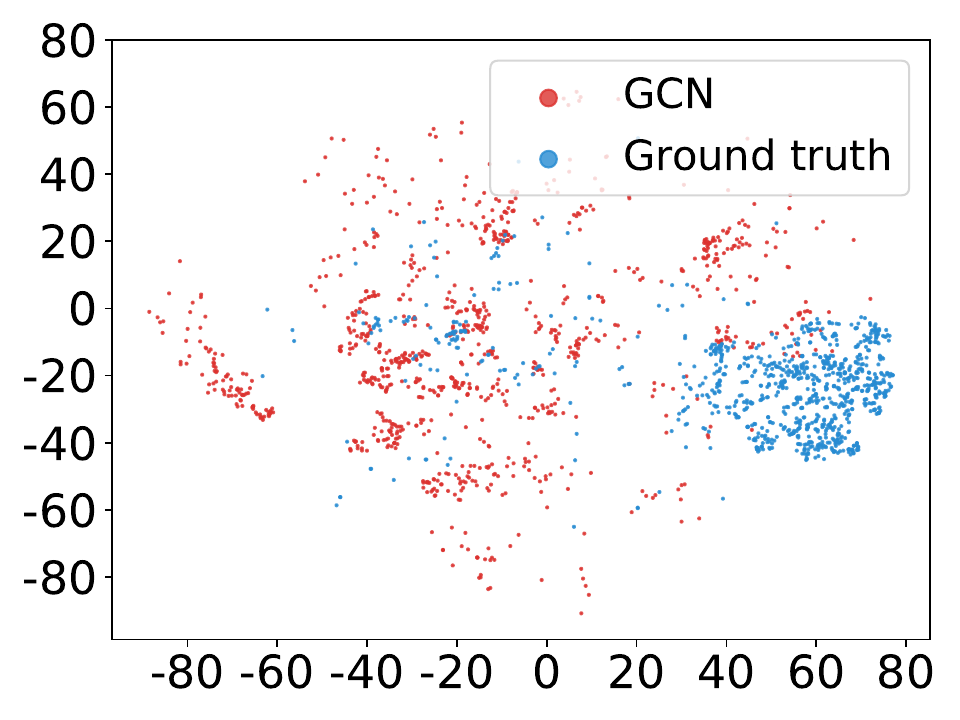}}
    \subfloat[DisC]{\label{fig:gap2}\includegraphics[width=0.2\linewidth]{
    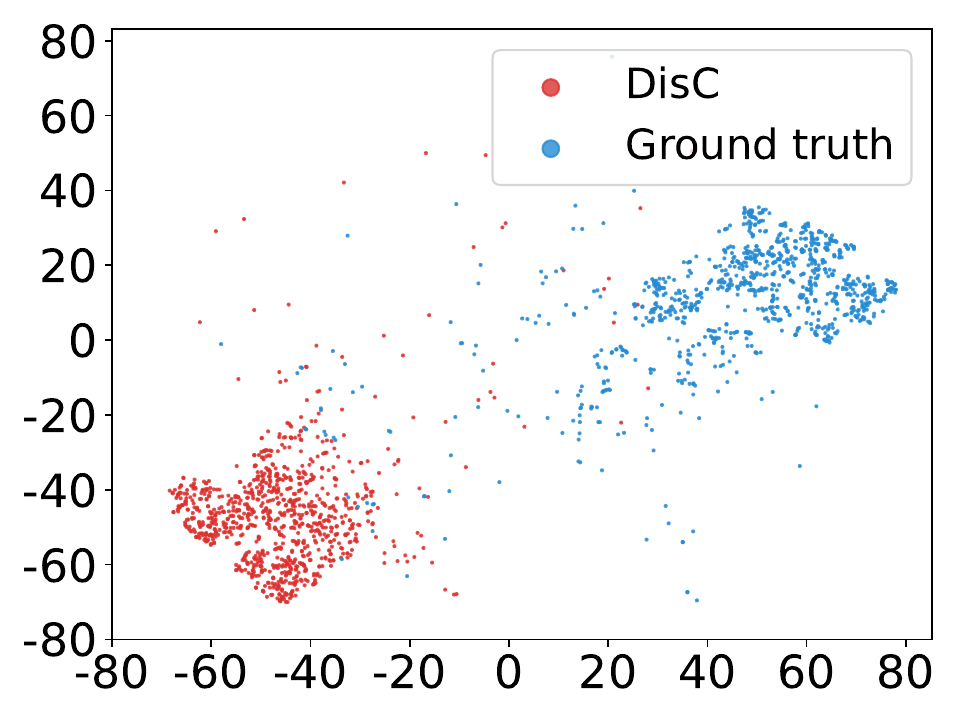}}
    \subfloat[CAL]{\label{fig:gap6}\includegraphics[width=0.2\linewidth]{
    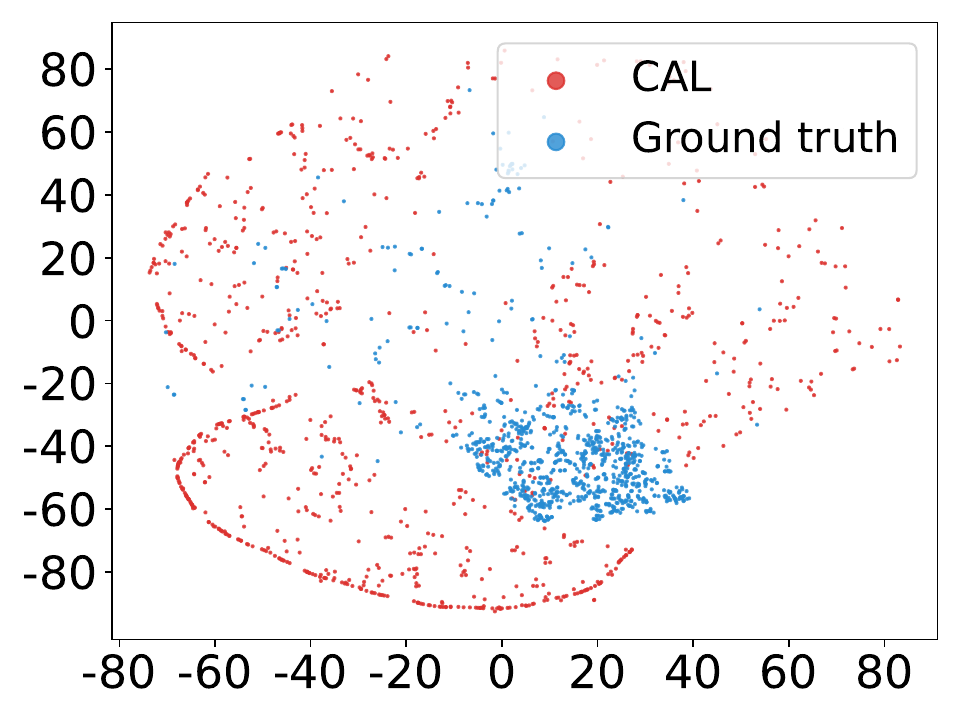}}
    \subfloat[GPro-1 step]{\label{fig:gap3}\includegraphics[width=0.2\linewidth]{
    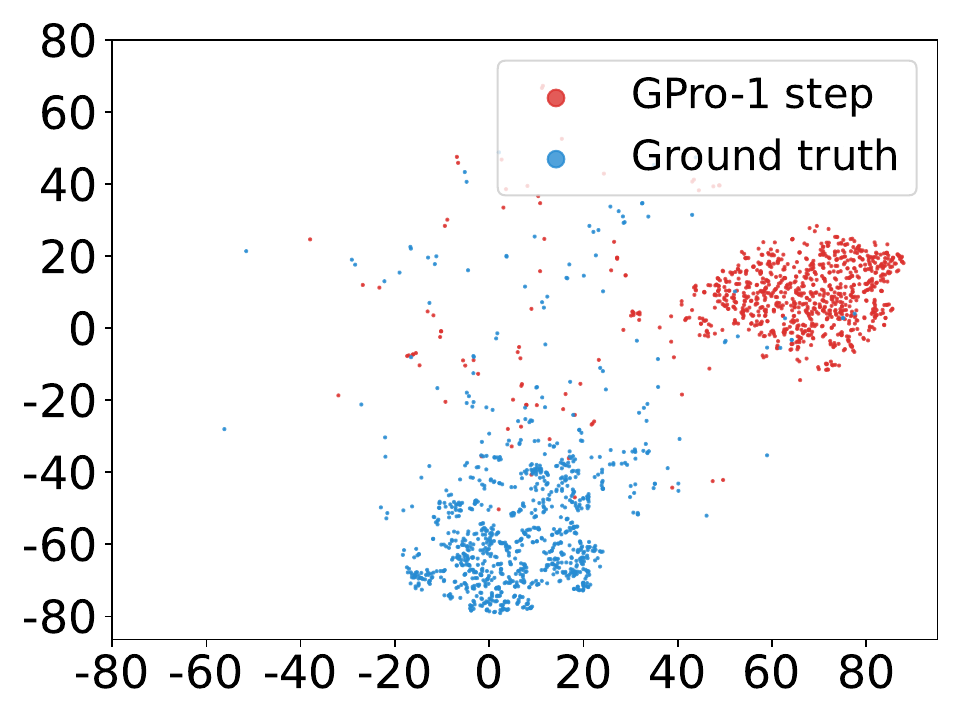}}
    \subfloat[GPro-4 step]{\label{fig:gap5}\includegraphics[width=0.2\linewidth]{
    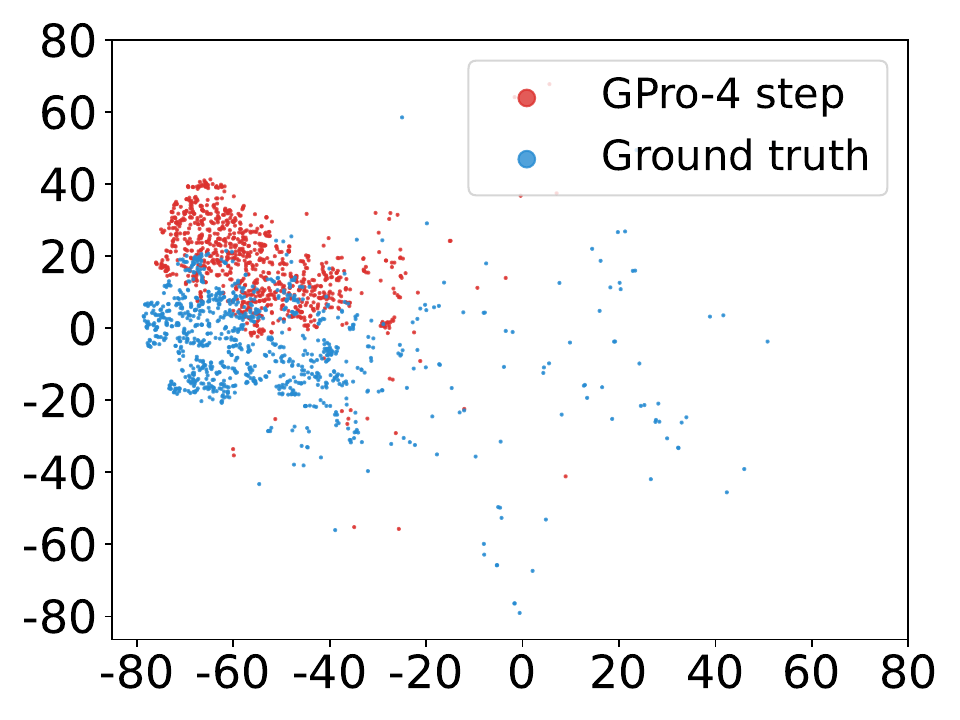}}\\
    \caption{TSNE visualization of sample features of class 0 generated by the model in the CMNIST-75sp dataset. There is generally a significant distribution gap between the features learned by existing methods (such as GCN, DisC and CAL) and the ground-truth causal features. GPro learns causal features that are closer to the ground-truth via progressive inference.}
    \label{fig:gap}    
\end{figure*}

To comprehensively verify the effectiveness of GPro, we compared 11 state-of-the-art algorithms and their variants.
Table~\ref{tab:main} shows the experimental results (\%) for the graph classification task in the three datasets. We report the mean accuracy and standard error. Bold indicates optimal and underline denotes suboptimal. On the basis of the experimental results, we can observe that GPro is optimal in 9 different dataset divisions. Specifically, the baselines developed based on ID are more likely to learn shortcut features from spurious correlations between non-causal parts and labels, resulting in performance that is typically inferior to OOD baselines. Compared to optimal ID-based baseline methods, GPro improves 22.81\%, 10.05\%, and 20.05\% on average in three datasets, respectively. When spurious correlations are more severe in the training set, that is, the bias is larger, the performance of the baseline developed based on ID degrades severely. GPro improves 13.91\%, 17.75\%, and 21.29\% on average over the ID-based design approach when the bias degree of the datasets is 0.8, 0.9, and 0.95, demonstrating that GPro has better debiasing causal learning ability. Algorithms designed for OOD often achieve better performance. 
Compared with state-of-the-art methods specially designed for OOD, our proposed model outperforms 4.91\% on average. In the case of datasets with more severe distribution shifts, the performance improvement could reach 6. 86\%. This further supports the observation in Figure~\ref{fig:gap} that existing methods are limited in disentangling the complex coupled associations between causal and non-causal substructures in graphs, resulting in the failure to extract ground truth causal features.

In summary, the experimental results demonstrate that GPro obtains state-of-the-art OOD generalization capability through a well-designed progressive inference process, counterfactual sample generation, and causal loss functions.

\begin{figure}
\centering
  \includegraphics[width=1\linewidth]{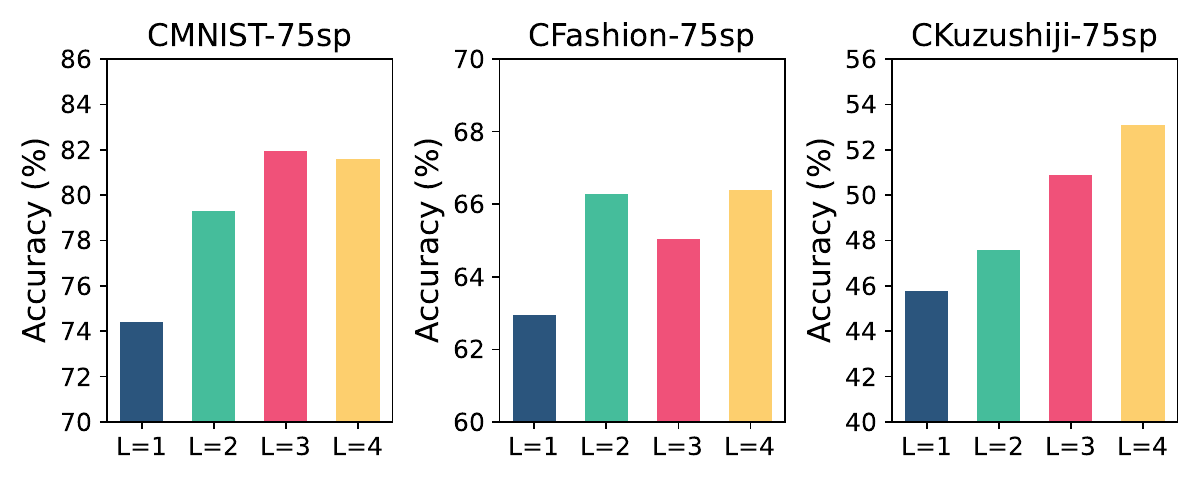}
 \caption{Quantitative sensitivity analysis of GPro for the number of progressive inference steps.}
    \label{fig:param_blocks}
\end{figure}

\begin{figure*}
\setlength{\belowcaptionskip}{-0.35cm}
\centering
\subfloat[GCN]{\label{fig:vis1}\includegraphics[width=.2\linewidth]{
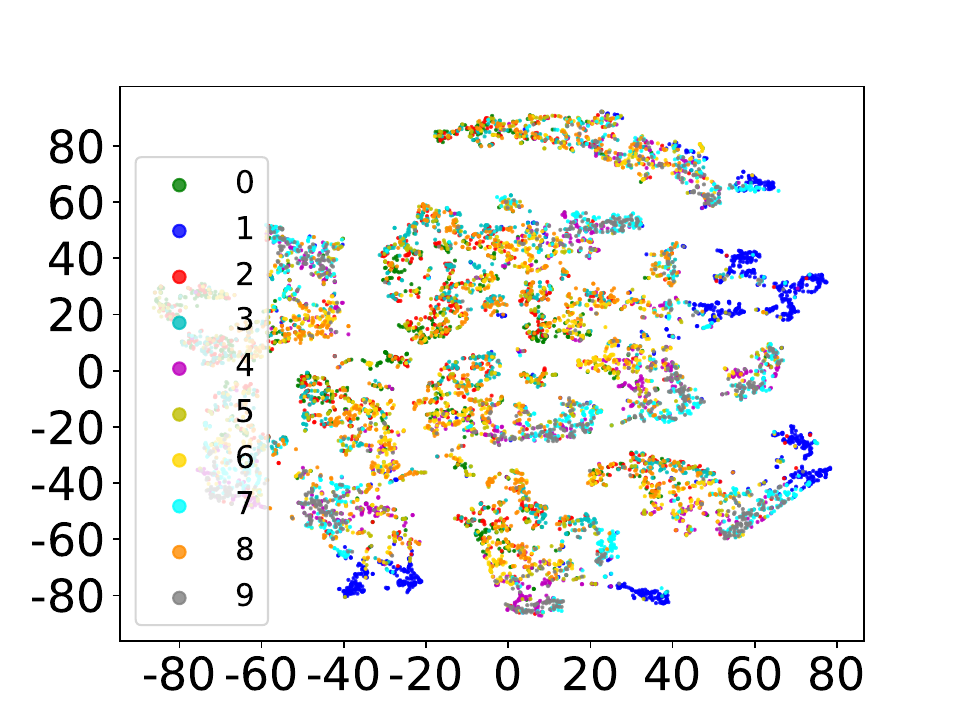}}
\subfloat[DisC]{\label{fig:vis2}\includegraphics[width=.2\linewidth]{
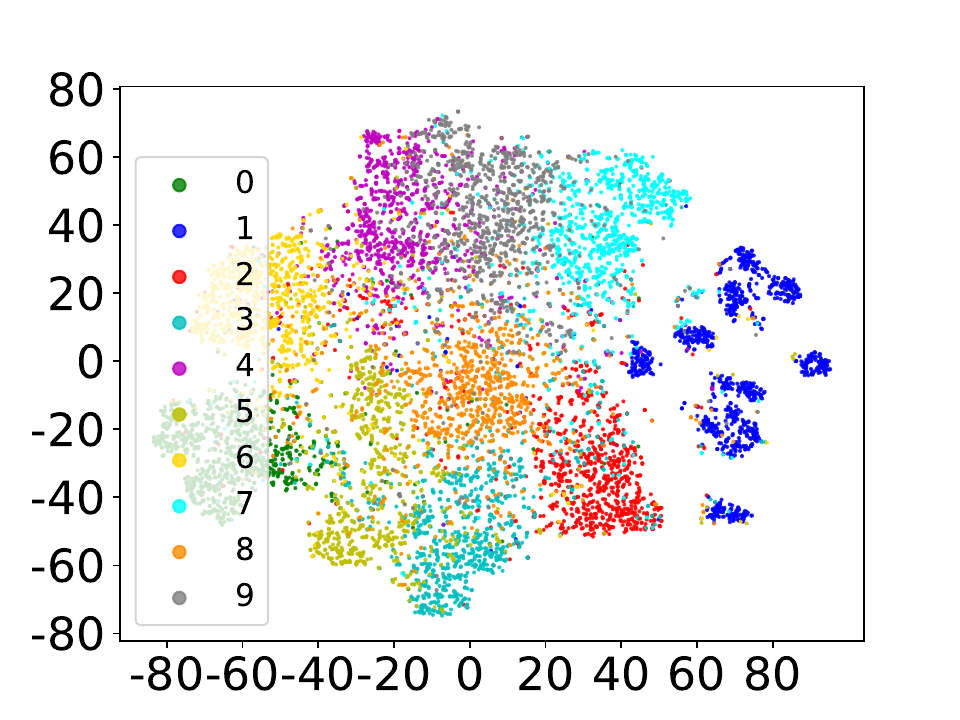}}
\subfloat[CAL]{\label{fig:vis6}\includegraphics[width=.2\linewidth]{
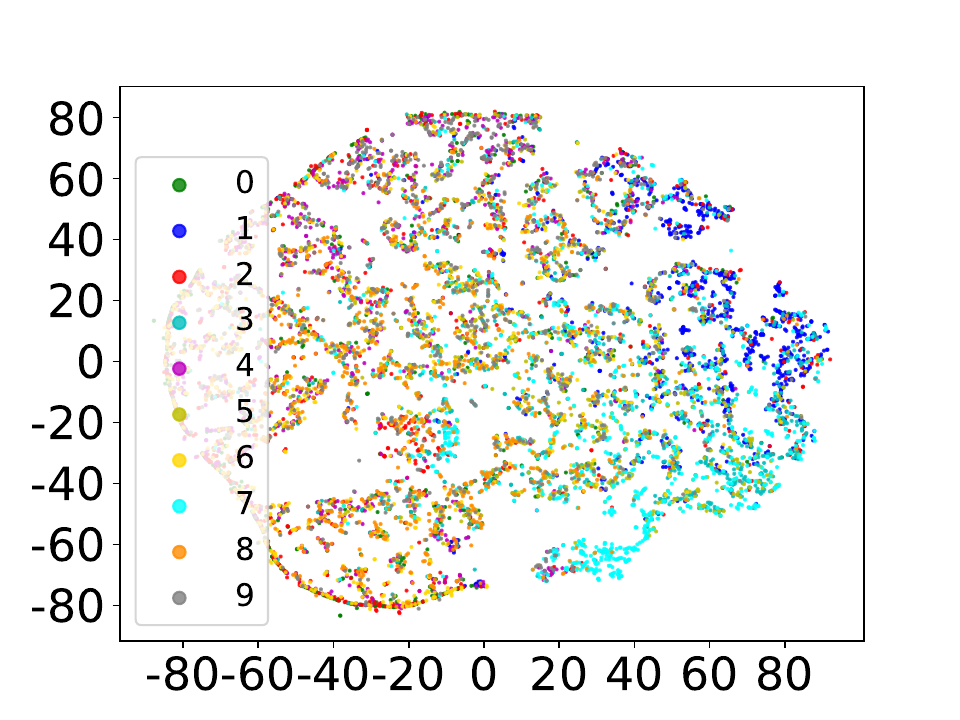}}
\subfloat[GPro-1 step]{\label{fig:vis3}\includegraphics[width=.2\linewidth]{
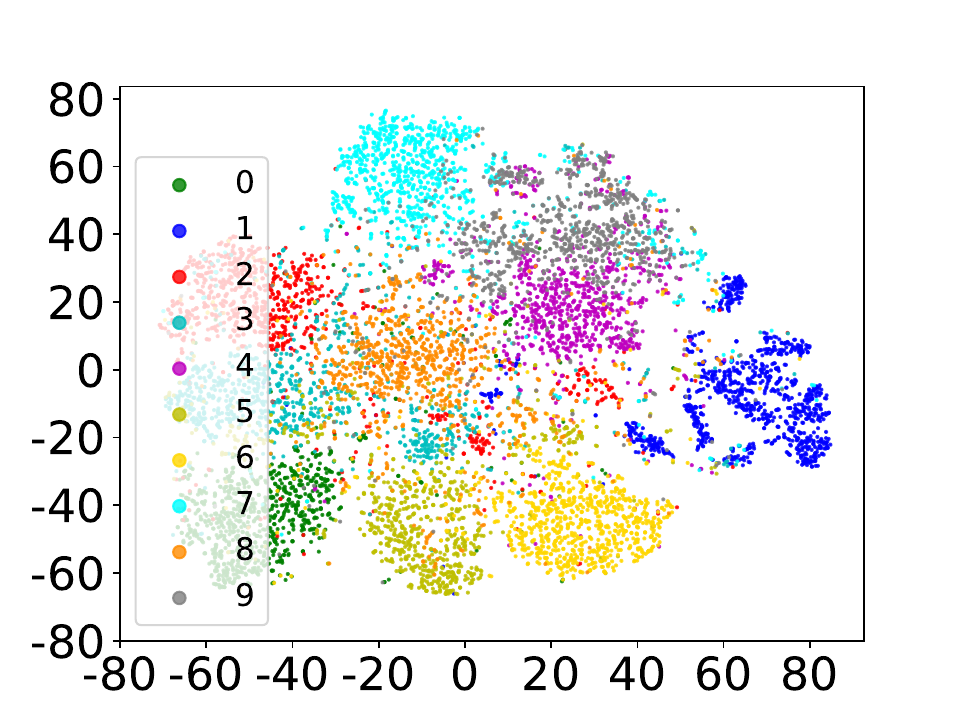}}
\subfloat[GPro-4 step]{\label{fig:vis5}\includegraphics[width=.2\linewidth]{
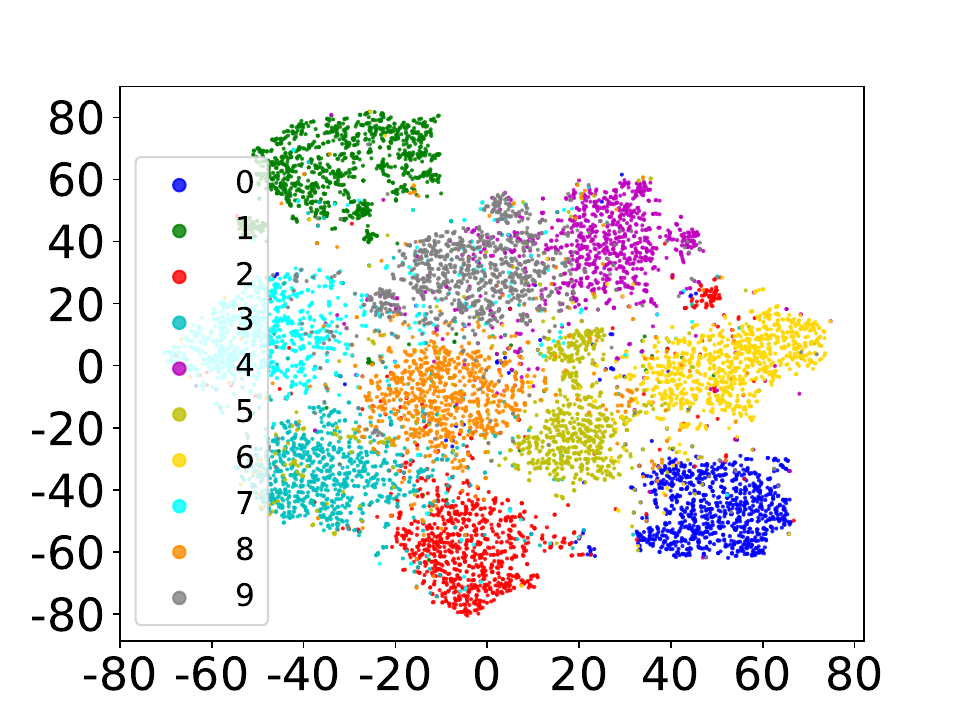}}\\	
\caption{TSNE visualization of the features learned by GCN, DisC, CAL, and GPro in the CMNIST-75sp dataset, where labels are marked by colors. The features learned through GPro show that the clusters within each category exhibit compactness while the distance between clusters is maximized.}
\label{fig:tsne}
\end{figure*}

\subsection{Effectiveness of Progressive inference}\label{sec:sec5.5}
This subsection evaluates progressive inference through quantitative and qualitative analyses.

\subsubsection{Quantitative Evaluation}
We quantitatively evaluated our model by comparing the accuracy (\%) across three challenging datasets: CMNIST-75sp-0.95, CFashion-75sp-0.95, and CKuzushiji-75sp-0.95. We assess the performance at 1, 2, 3, and 4 progressive inference steps, facilitated by stacking substructure context sampling blocks, with each block representing one step. Initial results, with a single inference step ($L=1$), show a 4.02\% improvement over the leading model, confirming the efficacy of GPro components. Performance typically improves as the number of inference layers increases. These findings suggest that more complex datasets require additional inference steps to achieve optimal performance, while simpler datasets are well served by 2 to 3 steps. 
This experiment illustrates the importance of a multi-step approach in graph causality analysis.

\subsubsection{Qualitative Visualization Evaluation}\label{sec:vis_pr}
We qualitatively evaluate the benefits of progressive inference in GPro using t-SNE visualization, as shown in Figure~\ref{fig:gap}. The visualization highlights significant gaps between the causal features learned by ID methods like GCN and OOD methods such as DisC and CAL, compared to ground-truth causal features, which leads to the predictions made by the existing methods being still unreliable.  GPro employs progressive inference to help bridge these gaps, with longer inference processes (4-step) yielding superior results compared to shorter inference processes (1-step). Moreover, we visualize the representations of all samples in the test set of CMNIST-75sp dataset learned by the above models. 
Figure~\ref{fig:vis1} shows that each cluster mixes multiple classes, indicating GCN tends to learn shortcut features (non-causal features) from spurious correlations between non-causal parts and labels, and fails to capture generalized causal features. 
Single-step methods like DisC, CAL, and GPro-1step inadequately distinguish class features, resulting in blurred cluster boundaries.
Conversely, employing longer inference steps results in tighter intra-class clusters and more distinct inter-class distances, showcasing the exceptional capability of progressive inference to capture causal features effectively.

\subsection{Ablation Studies}
To validate the validity of each component in GPro, we conduct ablation studies on CMNIST-75sp, CFashion-75sp, and CKuzushiji-75sp with all bias degrees of 0.9. 
Specifically, w/o $\mathbf{Z}_{n}^{per}$ and w/o $\mathbf{Z}_{n}^{smv}$ are designed to remove the counterfactual generation strategy of randomly permuting the non-causal representations and swapping the mean and variance between the non-causal representations, respectively.
W/o $\mathcal{L}_{\textrm{cou}}$, w/o $\mathcal{L}_{\textrm{scl}}$ and w/o $\mathcal{L}_{\textrm{con}}$ are the GPro variant models for removing $\mathcal{L}_{\textrm{cou}}$, $\mathcal{L}_{\textrm{scl}}$ , and $\mathcal{L}_{\textrm{con}}$ from the loss function Eq.~\eqref{eq:loss_function}, respectively. As shown in Table~\ref{tab:ablation}, we have the following observations: among the two counterfactual generation strategies, w/o $\mathbf{Z}_{n}^{per}$ performance decreases by 0.89\%, and w/o $\mathbf{Z}_{n}^{smv}$ performance decreases by 1.37\% on average in the three datasets. The effectiveness of two counterfactual generation strategies is demonstrated, while w/o $\mathbf{Z}_{n}^{smv}$ brings a more significant performance improvement. In addition, w/o $\mathcal{L}_{\textrm{cou}}$, w/o $\mathcal{L}_{\textrm{scl}}$, and w/o $\mathcal{L}_{\textrm{con}}$ show 7.15\%, 1.16\% and 0.71\% performance degradation on the three datasets, respectively. 
Removing $\mathcal{L}_{\textrm{cou}}$ significantly reduces performance across all datasets. Overall, omitting any component in GPro leads to performance degradation, underscoring the importance of each component.

\begin{table}
\renewcommand\arraystretch{1.36} 
\resizebox{1\hsize}{!}{
\centering
\begin{tabular}{cccc}
\toprule 
Method & CMNIST-75sp    & CFashion-75sp   & CKuzushiji-75sp   \\ \midrule
GPro   & $87.58_{\pm 0.36}$  & $70.57_{\pm 0.29}$ & $58.35_{\pm 0.63}$ \\ \hline
w/o $\mathbf{Z}_{n}^{per}$  & $86.84_{\pm 0.52} \left ( \downarrow_{0.74} \right )$ & $69.35_{\pm 0.86} \left ( \downarrow_{1.22} \right )$ & $57.63_{\pm 0.58} \left ( \downarrow_{0.72} \right )$ \\ 
w/o $\mathbf{Z}_{n}^{smv}$ & $86.71_{\pm 0.41} \left ( \downarrow_{0.87} \right )$ & $68.96_{\pm 0.38} \left ( \downarrow_{1.61} \right )$ & $56.73_{\pm 0.58} \left ( \downarrow_{1.62} \right )$ \\
w/o $\mathcal{L}_{\textrm{cou}}$  & $82.99_{\pm 0.65} \left ( \downarrow_{4.59} \right )$ & $63.82_{\pm 0.40} \left ( \downarrow_{6.75} \right )$ & $48.25_{\pm 0.77} \left ( \downarrow_{10.10} \right )$ \\ 
w/o $\mathcal{L}_{\textrm{scl}}$ & $86.25_{\pm 1.61} \left ( \downarrow_{1.33} \right )$  & $69.03_{\pm 0.32} \left ( \downarrow_{1.54} \right )$ & $57.73_{\pm 0.64} \left ( \downarrow_{0.62} \right )$ \\
w/o $\mathcal{L}_{\textrm{con}}$  & $86.67_{\pm 1.05} \left ( \downarrow_{0.91} \right )$ & $70.13_{\pm 0.33} \left ( \downarrow_{0.44} \right )$ & $57.57_{\pm 0.31} \left ( \downarrow_{0.78} \right )$  \\
\bottomrule
\end{tabular}}
\caption{Ablation study on different variants.}
\label{tab:ablation}
\end{table}

\section{Conclusion}
In this paper, we propose a novel approach to graph causal invariant learning via progressive inference perspective, called GPro. Specifically, we decompose the problem of identifying causal invariant parts of graphs into multiple intermediate inference steps, and extract causal features that are stable to distribution shifts through step-by-step inference. 
To make the progressive inference process better capture the causal invariant parts, we propose a novel feature augmentation method to generate counterfactual samples to enlarge the training distribution. Moreover, we propose a new supervised contrastive learning method to fully utilize supervised signals.
We conduct comprehensive experiments on three datasets. Compared with the state-of-the-art method, our proposed model outperforms 4.91\% on average. In the case of datasets with more severe distribution shifts, the performance improvement could be up to 6.86\%. The experimental results demonstrate that our proposed method is superior to the state-of-the-art methods.

\section{Acknowledgments}
This research was partially supported by the National Key Research and Development Project of China No. 2021ZD0110700, the Key Research and Development Project in Shaanxi Province No. 2022GXLH01-03, the National Science Foundation of China No. (62037001, 62250009, 62476215, 62302380), the China Postdoctoral Science Foundation No. 2023M742789, the Fundamental Scientific Research Funding No. (xzd012023061 and xpt012024003), and the Shaanxi Continuing Higher Education Teaching Reform Research Project No. 21XJZ014. Co-author Chen Chen consulted on this project on unpaid weekends for personal interests, and appreciated collaborators and family for their understanding.

\bibliography{aaai25}

\clearpage
\begin{center}
    \huge \textbf{Appendix} \label{sec:appendix}
\end{center}
\appendix
\renewcommand\thefigure{A\arabic{figure}}
\renewcommand\theequation{A\arabic{equation}}
\frenchspacing
\maketitle

\section{Notations}
We summarize the necessary notations used in Table~\ref{tab:ND}. 
We denote a graph $G=\left\{ \mathcal{V}, \mathcal{E} \right\}$ with the node set $\mathcal{V}$ and edge set $\mathcal{E}$. The node feature matrix $ \mathbf{X}=\left\{ \mathbf{x_{i}}|i\in \mathcal{V}\right\}\in \mathbb{R}^{\left| \mathcal{V}\right| \times F}$, where $F$ is node feature dimension and $\mathbf{x_{i}}= \mathbf{X}\left [ i,: \right ] $ is the $F$-dimensional attribute vector of node $v_i$. The adjacency matrix of graph $G$  is denoted as $\mathbf{A}\in \mathbb{R}^{\left| \mathcal{V}\right| \times \left| \mathcal{V}\right|}$, where $ \mathbf{A}\left [ i,j \right ]=1$ if edge $\left ( v_{i},v_{j} \right ) \in \mathcal{E}$, otherwise $ \mathbf{A}\left [ i,j \right ]=0$. The labels $Y=\left\{ y_{i}|i\in \mathcal{E}\right\}$.

\begin{table}
\centering
\renewcommand\arraystretch{1.2}
\setlength\tabcolsep{12pt}
  \caption{Notations and Descriptions.}
  \label{tab:ND}
\resizebox{1\hsize}{!}{
  \begin{tabular}{cc}
    \toprule
    Notations & Descriptions\\
    \midrule
    $G$ & Graph $G=\left\{ \mathbf{A}, \mathbf{X} \right\}$ \\
    $\mathbf{A}$ & The adjacency matrix of graph $G$ \\
    $\mathbf{X}$ & The node representation matrix of graph $G$ \\
    $\mathcal{V}$ & Node set of graph $G$ \\
    $\mathcal{E}$ & Edge set of graph $G$ \\
    $\mathcal{N}\left ( i \right )$ & The set of neighbors of node $i$ \\
    $\mathcal{G}_{train}, \mathcal{G}_{test}$ & The training and testing graph data \\
    $||$ & Concatenation operation \\
    $\left \lfloor \cdot \right \rfloor$ & Floor function \\
    $\alpha_{ij}$ & The edge-level attention score of edge~$\left ( i,j \right )$ \\
    $\mathbf{M}$ & The mask matrix \\
    $G_{c},G_{n}$ & Extracted causal and non-causal substructures \\
    $\mathbf{Z}_{c},\mathbf{Z}_{n}$ & Extracted causal and non-causal representation matrices \\
    $\Phi_{c}, \Phi_{n}$ & The causal classifier and non-causal classifier  \\
    $\tau$ & Temperature parameter in supervised contrastive loss \\
  \bottomrule
\end{tabular}}
\end{table}

\section{Algorithm}
The implementation details of our proposed GPro are presented in Algorithm~\ref{alg:GPro}.

\begin{algorithm}
\caption{GPro framework that learns graph causal invariance via progressive inference} 
\label{alg:GPro}
\begin{algorithmic}[1]
\SetKwInOut{Input}{Input}
\SetKwInOut{Output}{Output} 
\REQUIRE{Graph dataset $\mathcal{G}=\left\{ \left ( G_{i},Y_{i} \right )\right\}_{i=1}^{N}$, $f(\cdot)$, steps of progressive inference $L$} 
\ENSURE{The trained predictor $ f\left ( \cdot  \right ): \mathbb{G}\to \mathbb{Y}$}
\FOR {sampled minibatch $\mathcal{B}$ of graph dataset $\mathcal{G}$}
\FOR {$l\leftarrow 1$ \textbf{to} $L$} 
\STATE Generate representations $\mathbf{H}_{c}^{l},\mathbf{H}_{n}^{l}$ by Eq.~\eqref{eq:gnn}.
\STATE Calculate causal and non-causal attention score matrix $\mathbf{E}_{c}^{l}, \mathbf{E}_{n}^{l}$ by Eq.~\eqref{eq:attention-score}.
\STATE Calculate causal and non-causal mask matrices $\mathbf{M}_{c}^{l}, \mathbf{M}_{n}^{l}$ by Eq.~\eqref{eq:mask}.
\STATE Update the learned causal and non-causal subgraphs $\mathbf{A}_{c}^{l}, \mathbf{A}_{n}^{l}$ of the current inference step by Eq.~\eqref{eq:sep}.
\ENDFOR
\STATE Generate causal and non-causal representations $\mathbf{Z}_{c}$ and $\mathbf{Z}_{n}$ by Eq.~\eqref{eq:gcn1} and Eq.~\eqref{eq:gcn2}.
\STATE Concatenate $\mathbf{Z}=\left [ \mathbf{Z}_{c};\mathbf{Z}_{n} \right ]$.
\STATE Generate two counterfactual graph representations $\mathbf{Z}^{per}=\left [ \mathbf{Z}_{c};\mathbf{Z}_{n}^{per} \right ]$ and $\mathbf{Z}^{smv}=\left [ \mathbf{Z}_{c};\mathbf{Z}_{n}^{smv} \right ]$ by Eq.~\eqref{eq:permute} and Eq.~\eqref{eq:swap}.
\STATE Calculate the total loss $\mathcal{L}$ by Eq.~\eqref{eq:loss_function}.
\STATE Update model parameters to minimize $\mathcal{L}$.
\ENDFOR
\end{algorithmic}
\end{algorithm}

\section{Baselines Detail}

\begin{table*}
  \caption{Statistics of Biased Graph Classification Datasets.}
  \centering
  \resizebox{0.7\textwidth}{!}{
  \begin{tabular}{cccccc}
    \toprule
    \textbf{Dataset} & \textbf{\#Graphs (train/val/test)} & \textbf{\#Avg. Nodes}  & \textbf{\#Avg. Edges} & \textbf{\#Classes}\\
    \midrule
    CMNIST-75sp & 10K/5K/10K & 61.09 & 488.78 & 10\\
    CFashion-75sp &10K/5K/10K & 61.03 & 488.26 & 10\\
    CKuzushiji-75sp & 10K/5K/10K & 52.81 & 422.47 & 10 \\
    \bottomrule
  \end{tabular}}
  \label{tab:statistics}
\end{table*}

\subsubsection{Baselines}
\textbf{In-Distribution Methods:} 
GCN~\cite{kipf2016semi}, GIN~\cite{xu2018powerful}, GCNII~\cite{chen2020simple}, FactorGCN~\cite{yang2020factorizable}, and DiffPool~\cite{ying2018hierarchical}.

\begin{itemize}
    \item GCN~\cite{kipf2016semi}: It is a foundational graph neural network model that introduces a semi-supervised learning architecture for graph data and has inspired many variants. It leverages convolutional operations to aggregate node features from neighborhoods, incorporating the graph’s topological structure.
    
    \item GIN~\cite{xu2018powerful}: It enhances structural information capture in graphs, optimizing for graph isomorphism testing by employing parameterized aggregation functions that simulate the Weisfeiler-Lehman test, ensuring strong discriminative power through theoretical guarantees.
    
    \item GCNII~\cite{chen2020simple}: It extends GCN~\cite{kipf2016semi} by incorporating residual and identity mapping to mitigate over-smoothing in deep GCNs, enabling the preservation of node information and graph structural details across layers. 
    
    \item FactorGCN~\cite{yang2020factorizable}: It factorizes the convolution into independently computable tasks, reducing parameter count and complexity for scalability on large graphs while capturing latent relations through disentangled aggregation.
    
    \item DiffPool~\cite{ying2018hierarchical}: It is a hierarchical representation learning approach that employs a differentiable pooling module to generate multi-scale node and subgraph representations, enabling the integration of GNNs with hierarchical pooling and capturing complex structural patterns at various scales through an end-to-end trainable framework.

\end{itemize}

\textbf{Out-of-Distribution Methods:} LDD~\cite{lee2021learning}, StableGNN~\cite{fan2023generalizing}, CAL~\cite{sui2022causal}, DisC~\cite{fan2022debiasing}, CIGA~\cite{chen2022learning} and GALA~\cite{chen2023does}.

\begin{itemize}
    \item LDD~\cite{lee2021learning}: It learns debiased representations by disentangling feature augmentations, separating them into distinct subspaces to reduce bias and correlation, leading to improved fairness and generalization in scenarios with biased or imbalanced data.
    
    \item StableGNN~\cite{fan2023generalizing}: It enhances the generalization of GNNs to out-of-distribution graphs by introducing a regularization term during training, learning more generalized feature representations robust to distribution shifts. \looseness=-1
    
    \item CAL~\cite{sui2022causal}: It improves interpretability and generalizability of graph classification by incorporating causal attention and debiasing, enabling the model to learn causal relationships and reduce bias.
    
    \item DisC~\cite{fan2022debiasing}: It learns disentangled causal substructures within graphs to identify and mitigate biases, using a multi-stage framework that combines causal discovery, disentanglement, and GNN modules for robust and generalizable representations.
     
    \item CIGA~\cite{chen2022learning}: It learns causally invariant representations by capturing causal relationships within graphs, enabling generalization to unseen graphs by characterizing distribution shifts with causal models to extract informative subgraphs and maximally preserving invariant intra-class information.
    
    \item GALA~\cite{chen2023does}: It is an enhancement of CIGA~\cite{chen2022learning} and learns more robust and generalizable graph representations through environment augmentation, employing an environment assistant model to detect and mitigate spurious correlations.
    
\end{itemize}

\begin{table*}[ht]\large
\caption{Experimental results (\%) for the graph classification task on unseen unbiased testing sets. We report the mean accuracy and standard error. Bold indicates the optimal and underline indicates the suboptimal.}
\label{tab:unseen}
\renewcommand\arraystretch{1.4}
\resizebox{1\textwidth}{!}{

 \centering
\begin{tabular}{c|ccc|ccc|ccc}
\hline \hline
\multicolumn{1}{c|}{Dataset}                                         & \multicolumn{3}{c|}{CMNIST-75sp}                                                                                                                        & \multicolumn{3}{c}{CFashion-75sp}     & \multicolumn{3}{|c}{CKuzushiji-75sp}                                                                                        \\ 
\multicolumn{1}{c|}{Bias}                                            & \multicolumn{1}{c}{0.8}               & \multicolumn{1}{c}{0.9}                & \multicolumn{1}{c|}{0.95} & \multicolumn{1}{c}{0.8}                             & \multicolumn{1}{c}{0.9}                 & \multicolumn{1}{c|}{0.95} & \multicolumn{1}{c}{0.8}                             & \multicolumn{1}{c}{0.9}                 & \multicolumn{1}{c}{0.95} \\ \hline

GCN~\cite{kipf2016semi}     & $36.88_{\pm 5.16}$  & $23.07_{\pm 4.07}$ & $11.88_{\pm 0.33}$  & $59.33_{\pm 0.55}$ & $53.65_{\pm 0.47}$ & $45.60_{\pm 1.06}$ & $36.35_{\pm 0.48}$ & $27.88_{\pm 0.94}$ & $19.95_{\pm 0.67}$  \\ 

GIN~\cite{xu2018powerful}     & $48.93_{\pm 2.99}$ & $34.95_{\pm 0.86}$ & $14.53_{\pm 0.97}$  & $58.88_{\pm 0.57}$  & $53.80_{\pm 0.52}$ & $48.43_{\pm 0.69}$ & $39.25_{\pm 0.57}$  & $30.75_{\pm 1.45}$ & $22.35_{\pm 0.86}$\\ 

GCNII~\cite{chen2020simple}     & $53.50_{\pm 6.23}$ & $45.52_{\pm 2.26}$ & $32.60_{\pm 5.66}$  & $58.85_{\pm 1.89}$ & $53.98_{\pm 0.85}$ & $46.97_{\pm 1.38}$ & $39.93_{\pm 0.88}$ & $30.33_{\pm 1.17}$ & $23.09_{\pm 1.83}$ \\  \hline

$\text{CAL}_{GCN}$~\cite{sui2022causal}   &  $77.27_{\pm 0.81}$  & $69.41_{\pm 0.52}$ &    $51.23_{\pm 1.52}$     & $68.08_{\pm 0.28}$  & $61.85_{\pm 0.58}$  & $55.63_{\pm 0.66}$ & $52.92_{\pm 0.75}$ & $42.40_{\pm 0.93}$ &  $32.20_{\pm 0.63}$\\ 
$\text{CAL}_{GIN}$~\cite{sui2022causal}     & $74.85_{\pm 1.04}$   & $65.11_{\pm 0.75}$  & $43.42_{\pm 1.89}$        & $66.31_{\pm 0.16}$ & $61.40_{\pm 0.36}$  & $55.29_{\pm 0.41}$ & $52.79_{\pm 0.65}$  & $41.17_{\pm 0.35}$ & $29.56_{\pm 0.99}$ \\ 
$\text{CAL}_{GAT}$~\cite{sui2022causal}     & $\mathbf{89.41}_{\pm 0.57}$ & $\underline{83.66_{\pm 0.38}}$ & $\underline{71.41_{\pm 1.44}}$        &  $\underline{71.78_{\pm 0.47}}$  & $67.51_{\pm 0.59}$ & $59.74_{\pm 0.86}$ & $\underline{66.63_{\pm 0.23}}$  & $\underline{56.61_{\pm 1.21}}$ & $\underline{42.73_{\pm 0.60}}$ \\ 
$\text{DisC}_{GCN}$~\cite{fan2022debiasing}     & $82.73_{\pm 1.31}$ & $77.70_{\pm 0.87}$ & $65.48_{\pm 0.76}$   & $67.90_{\pm 1.45}$ & $\underline{68.28_{\pm 0.18}}$ & $\underline{63.77_{\pm 1.37}}$   & $57.80_{\pm 2.38}$ & $51.60_{\pm 0.41}$ & $41.60_{\pm 3.94}$  \\ 

$\text{DisC}_{GIN}$~\cite{fan2022debiasing}     & $77.80_{\pm 1.33}$ & $73.00_{\pm 0.61}$ & $58.80_{\pm 1.66}$  & $67.15_{\pm 0.79}$  & $59.98_{\pm 0.62}$ & $51.70_{\pm 0.34}$  & $55.47_{\pm 0.98}$  & $43.20{\pm 1.36}$ & $31.33_{\pm 1.71}$ \\ 

$\text{DisC}_{GCNII}$~\cite{fan2022debiasing}     & $79.65_{\pm 2.13}$  & $76.63_{\pm 1.38}$ & $60.00_{\pm 5.66}$   & $60.50_{\pm 2.77}$& $63.05_{\pm 2.25}$ & $61.78_{\pm 1.60}$  & $56.23_{\pm 3.45}$ & $49.10_{\pm 2.05}$ & $41.05_{\pm 0.11}$\\

$\text{CIGA}$~\cite{chen2022learning}     & $62.66_{\pm 2.81}$ & $53.36_{\pm 2.70}$ & $37.17_{\pm 4.84}$        & $60.12_{\pm 2.73}$  & $55.80_{\pm 1.81}$ & $48.67_{\pm 2.76}$ & $47.71_{\pm 2.64}$  & $37.92_{\pm 1.34}$& $25.61_{\pm 1.82}$ \\
$\text{GALA}$~\cite{chen2023does}     & $73.16_{\pm 2.64}$ & $66.06_{\pm 3.17}$ & $35.54_{\pm 6.22}$        & $61.16_{\pm 0.71}$  & $57.83_{\pm 1.14}$ & $50.04_{\pm 1.22}$ & $48.36_{\pm 3.27}$  & $33.36_{\pm 3.41}$& $25.62_{\pm 0.85}$ \\ \hline

$\text{GPro}$     & $\underline{88.78_{\pm 0.93}}$ & $\mathbf{87.82}_{\pm 0.25}$ & $\mathbf{79.54}_{\pm 1.02}$        & $\mathbf{75.69}_{\pm 0.42}$  & $\mathbf{71.58}_{\pm 0.25}$ & $\mathbf{66.00}_{\pm 0.65}$ & $\mathbf{66.78}_{\pm 0.75}$  & $\mathbf{60.52}_{\pm 0.53}$& $\mathbf{49.29}_{\pm 0.37}$ \\
\hline \hline

\end{tabular}}
\end{table*}

\section{More Experiments}
\subsection{Robustness Analysis on Unseen Bias}
To further investigate the robustness and generalization of GPro, we report the results of the model on the unseen unbiased testing sets in Table~\ref{tab:unseen}, i.e., the predefined bias sets (non-causal parts) in the training sets and testing sets are disjoint. As shown in Table~\ref{tab:unseen}, GPro still outperforms the other models in the unseen unbiased testing sets, where GPro is optimal in 8 metrics and suboptimal in 1 metric. The performance of ID-based models further degrades in unseen bias scenarios, where the performance of GCN decreases on average by 7.02\%, 3.31\%, and 1.11\% on the three datasets. GIN decreases by 4.05\%, 3.64\%, and 0.25\%, respectively. GCNII decreases by 12.25\%, 6.88\%, and 5.67\% on average. This again demonstrates that the ID-based model learns shortcut features from spurious correlations rather than true causal features. However, GPro in the unseen unbiased testing sets, compared to Table~\ref{tab:main} performance, even improves by 0.35\%, 0.86\%, and 1.41\% in the three datasets, respectively. Furthermore, our proposed model outperforms the state-of-the-art method by 3.52\% on average. Meanwhile, GPro improves 8.13\%, 2.23\%, and 6.56\% in three datasets with a bias degree of 0.95 compared to the suboptimal baseline method designed specifically for OOD. This proves that GPro has stronger robustness and generalization ability.

\subsection{Flexibility Studies}
We integrate GPro with different GNN models, namely GCN, GIN, and GraphSAGE, to validate the flexibility of GPro on three datasets with a bias degree of 0.9. Table~\ref{tab:flexibility} summarizes the experimental results. We observe that GPro with different GNN encoders is still highly competitive in all three datasets. When the GNN encoder employs GraphSAGE, it improves 4.03\%, 4.35\%, and 3.57\%, respectively, compared to the best baseline existing. In addition, the GNN encoder with GIN improves 19. 76\%, 6. 59\%, 17. 29\% on average in the three datasets compared to $\text{LDD}_{GIN}$, $\text{CAL}_{GIN}$, and $\text{DisC}_{GIN}$, respectively. The experimental results showcase the remarkable flexibility of our proposed GPro framework, as it seamlessly integrates with various GNN encoders and consistently achieves superior performance. \looseness=-1

\begin{table}
\centering
\renewcommand\arraystretch{1.4}
\caption{The flexibility is verified by the effect of different GPro variants on the accuracy (\%).}
\resizebox{1\hsize}{!}{
\begin{tabular}{cccc}
\toprule
\small{GNN encoder}  & \small{CMNIST-75sp}    & \small{CFashion-75sp}   & \small{CKuzushiji-75sp}   \\ \midrule
GraphSAGE   & $85.60_{\pm 0.76}$  & $70.57_{\pm 0.39}$ & $55.57_{\pm 1.17}$ \\
GIN   & $83.36_{\pm 1.69}$  & $65.53_{\pm 0.38}$ & $53.66_{\pm 0.78}$ \\
GCN (default)   & $87.58_{\pm 0.36}$  & $70.57_{\pm 0.29}$ & $58.35_{\pm 0.63}$ \\
\bottomrule
\end{tabular}
\label{tab:flexibility}}
\end{table}

\begin{figure*}
	\centering
	\subfloat[Trade-off parameter $\lambda_{1}$]{\label{fig:param1}\includegraphics[width=.25\linewidth]{
    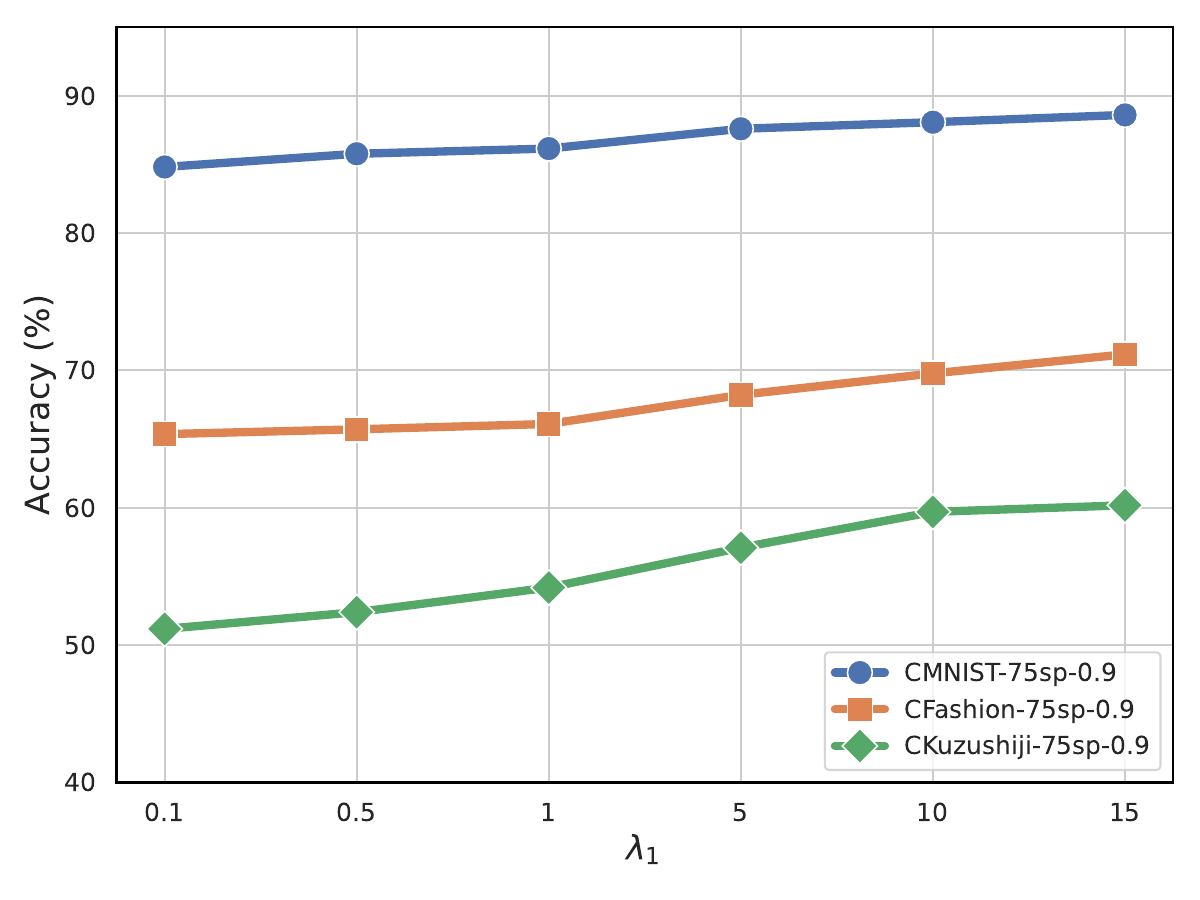}}
	\subfloat[Trade-off parameter $\lambda_{2}$]{\label{fig:param2}\includegraphics[width=.25\linewidth]{
    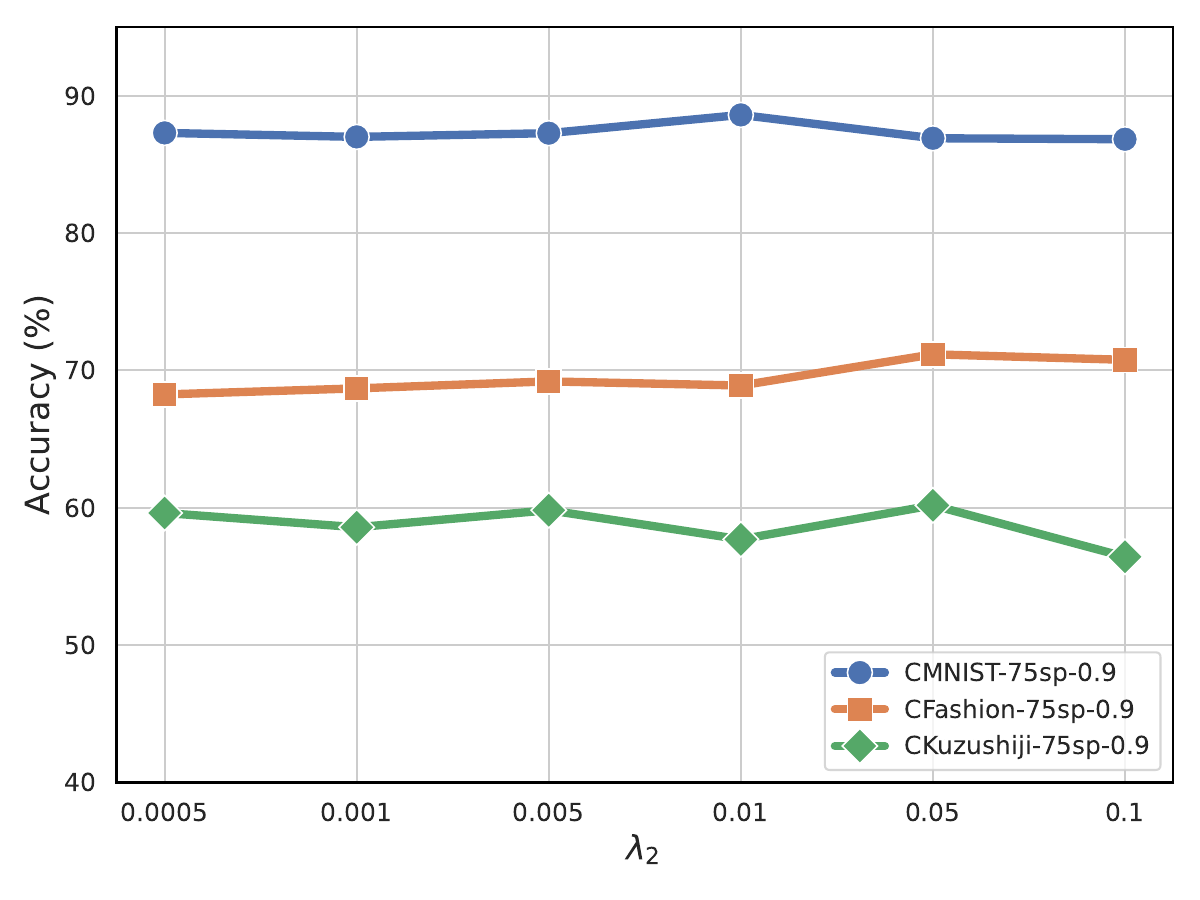}}
    \subfloat[Trade-off parameter $\lambda_{3}$]{\label{fig:param3}\includegraphics[width=.25\linewidth]{
    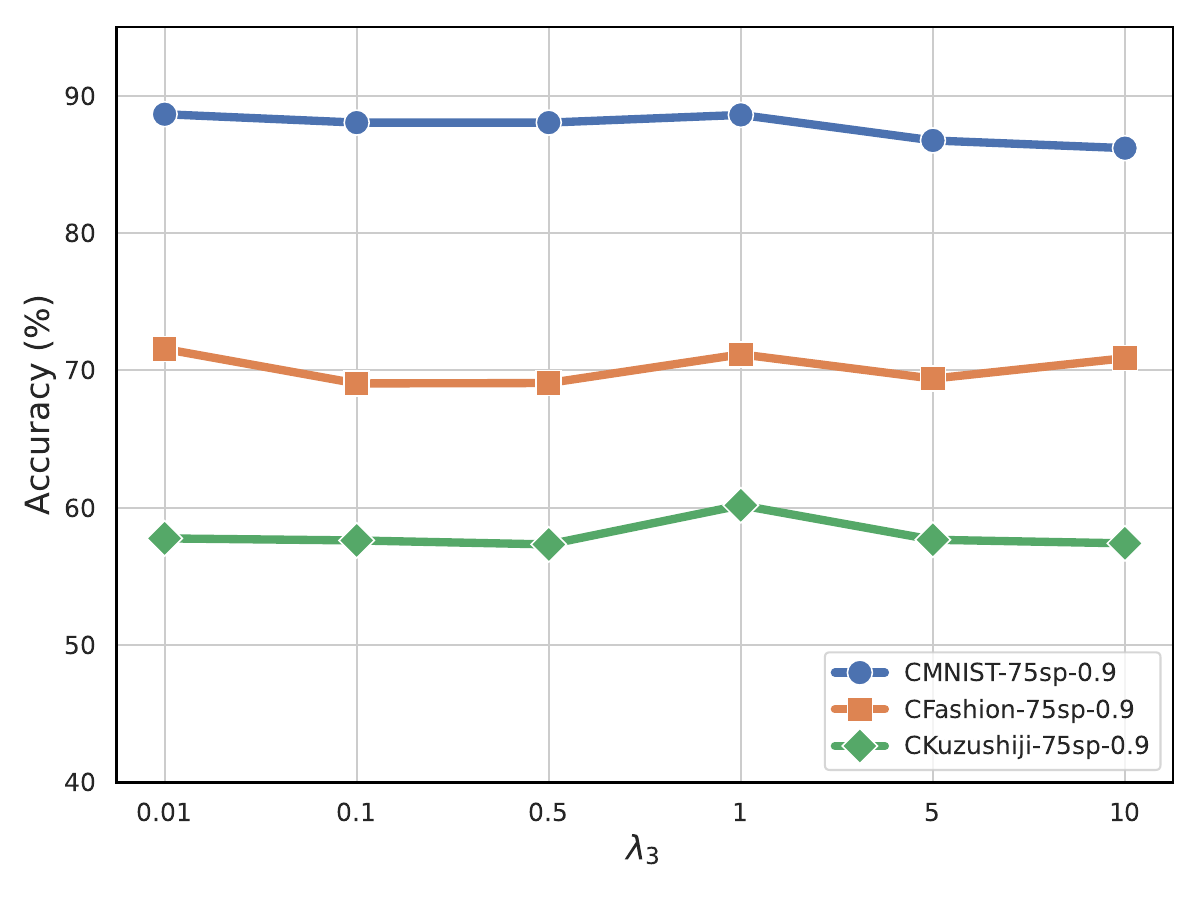}}    
    \subfloat[Temperature parameter $\tau$]{\label{fig:param4}\includegraphics[width=.25\linewidth]{
    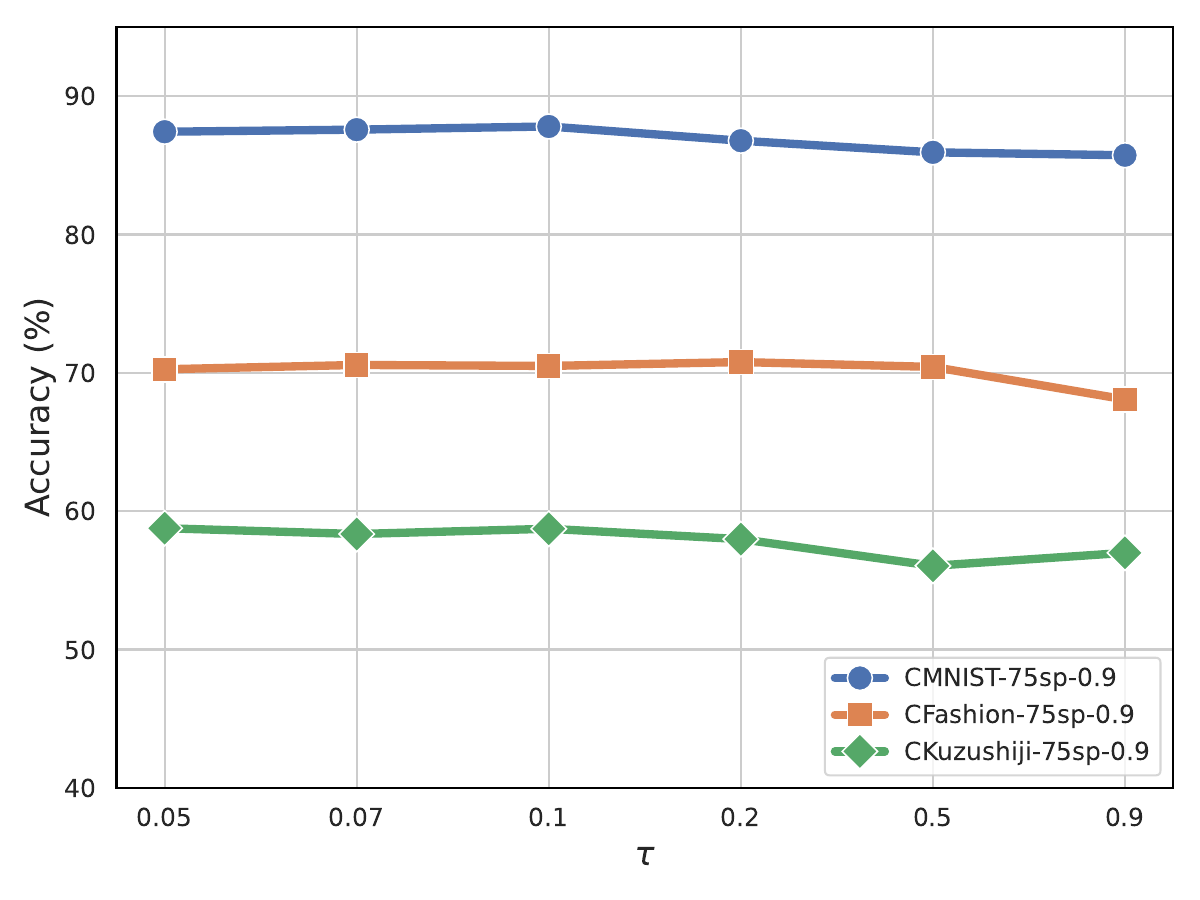}}\\	
    \caption{Parameter sensitivity of GPro for loss function coefficients $\lambda_{1}$, $\lambda_{2}$, $\lambda_{3}$, and temperature parameter $\tau$ on three datasets w.r.t. accuracy (\%), respectively.}
    \label{fig:param_loss}
\end{figure*}

\subsection{Hyperparameters Sensitivity}
In this subsection, we investigate our model GPro using different hyperparameters settings, i.e., trade-off parameter $\lambda_{1}$, $\lambda_{2}$, and $\lambda_{3}$ in the loss function Eq.~\eqref{eq:loss_function} and temperature parameter $\tau$ in Eq.~\eqref{eq:loss_3}. We report the accuracy (\%) results for the three datasets CMNIST-75sp-0.9, CFashion-75sp-0.9, and CKuzushiji-75sp-0.9 with different parameters. 

\textbf{Effect of trade-off parameter $\lambda_{1}$}\quad 
We investigate the effect of varying the value of $\lambda_{1}$ of loss $\mathcal{L}_{\textrm{cou}}$ on performance. The experimental results are shown in Figure~\ref{fig:param1}. We set $\lambda_{1}$ to 0.1, 0.5, 1, 5, 10, and 15 respectively for experiments on three datasets. The results are shown in Figure~\ref{fig:param1}, all three curves exhibit an overall upward trend in terms of accuracy. In particular, the larger $\lambda_{1}$ is more significant for performance improvement in the more difficult datasets.

\textbf{Effect of trade-off parameter $\lambda_{2}$}\quad In Figure~\ref{fig:param2}, we present the evaluation results for different values of $\lambda_{2}$ in Eq.~\eqref{eq:loss_function}. We set $\lambda_{2}$ to 0.0005, 0.001, 0.005, 0.01, 0.05, and 0.1 in all datasets. We observe that regardless of the value of lam2, increasing the supervised contrast loss leads to better performance than the existing baseline, similar conclusions are drawn with the ablation experiments. Meanwhile, GPro exhibits robustness to different $\lambda_{2}$ over all datasets.

\textbf{Effect of trade-off parameter $\lambda_{3}$}\quad As depicted in Figure~\ref{fig:param3}, we visualize the effect of the $\lambda_{3}$ on the performance of GPro. We set $\lambda_{3}$ to 0.01, 0.1, 0.5, 1, 5, and 10 for the experiments, respectively. When $\lambda_{3}$ exceeds 1 and $\mathcal{L}_{\textrm{con}}$ loss is too large, performance will decrease slightly. Similar to $\lambda_{2}$, GPro achieves the SOTA effect under all $\lambda_{3}$ value settings, and the performance of GPro in all three datasets is very robust. \looseness=-1

\textbf{Effect of temperature parameter $\tau$}\quad 
We investigate the effect of the hyperparameter sensitivity analysis of the temperature parameter in Eq.~\eqref{eq:loss_3} of the supervised contrastive loss we proposed, which is often used to control the sharpness of the similarity scores or logits produced by the contrastive loss function. We set to 0.07 as used by most papers by default, here we set it to 0.05, 0.07, 0.1, 0.2, 0.5, and 0.9. From the results, our model exhibits remarkable robustness. The performance decreases when the temperature coefficient is 0.5 or 0.9. However, it is worth noting that our method still outperforms baselines regardless of the hyperparameter choice. \looseness=-1

\begin{table}
\centering
\renewcommand\arraystretch{1.4} 
\caption{Total training time (seconds) on various datasets.}
\begin{tabular}{cccc}
\toprule
\small{Model}  & \small{CMNIST-75sp}    & \small{CFashion-75sp}   & \small{CKuzushiji-75sp}   \\ \midrule
CIGA   & $3,746s$  & $3,574s$  & $3,672s$ \\ %
GALA   & $3,864s$  & $4,150s$ & $3,786s$ \\
GPro   & $5,690s$  & $6,320s$ & $5,182s$ \\
\bottomrule
\end{tabular}
\label{tab:time}
\end{table}

\begin{table*}
\centering
\caption{Performance comparison across different datasets and metrics.}
\label{tab:results}
\begin{tabular}{cccccccc}
\toprule
\textbf{Datasets} & \textbf{EC50-Assay} & \textbf{EC50-Sca} & \textbf{EC50-Size} & \textbf{Ki-Assay} & \textbf{Ki-Sca} & \textbf{Ki-Size} & \textbf{Avg.} \\
\midrule
CAL   & 75.10 & 64.79 & 63.38 & 75.22 & 71.08 & 72.93 & 70.42 \\
DisC  & 61.94 & 54.10 & 57.64 & 54.12 & 55.35 & 50.83 & 55.66 \\
GALA  & 77.56 & 66.28 & 64.25 & 77.92 & 73.17 & 77.40 & 72.76 \\
GPro  & 86.68 & 73.03 & 71.38 & 92.84 & 91.92 & 92.92 & 84.80 \\
\bottomrule
\end{tabular}
\end{table*}

\subsection{Time Complexity Studies}
For simplicity of analysis, we assume that the encoder forward propagation complexity of the existing single-step method is $O\left ( f \right ) $, and the complexity of the backward propagation is $O\left ( b \right ) $. $L$ is the number of progressive learning steps in our work, then the corresponding encoder forward propagation complexity and backward propagation complexity of GPro are $O\left ( Lf \right ) $ and $O\left ( Lb \right ) $, respectively. The above experiments show that $L$ is usually not a very large integer, e.g., in this paper $L\leq 4$. Therefore, the time complexity of the model encoder does not increase significantly. The time complexity of the loss function $\mathcal{L}_{\textrm{scl}}$ and $\mathcal{L}_{\textrm{con}}$ are $O\left ( \left| \mathcal{G}\right|d^{2} \right ) $ and $O\left ( \left| \mathcal{G}\right|d \right )$, where $\left| \mathcal{G}\right|$ denotes the number of graphs, and $d$ is the dimension of the representations. In contrast, the time complexity of the encoder message passing mechanism of the single-step model can be approximated as $O\left ( \left| \mathcal{G}\right|\left| V\right|d^{2} + \left| \mathcal{G}\right|\left| E\right|d\right)$~\cite{li2022learning}, where $\left| V\right|$ and $\left| E\right|$ denote the number of nodes and edges in each graph. Compared to message passing, the introduction of the loss function incurs a significantly smaller increase in time complexity. Hence, the time complexity of GPro is comparable to previous works, demonstrating its promising efficiency.

We also compare the running time of GPro with some baselines quantitatively. The total training time (seconds) is shown in Table~\ref{tab:time}. The total training time for GPro increases by 1.56 and 1.46 times than CIGA and GALA, respectively. This observation aligns with the analysis of time complexity, providing evidence of a limited overhead increase.

\subsection{DrugOOD benchmark}
To cover more realistic situations, we conduct additional experiments on six DrugOOD benchmark datasets from GALA~\cite{chen2023does}. DrugOOD benchmark datasets focus on drug affinity prediction. As shown in the table below, GPro achieved the best results, outperforming baselines by over 12\%, which further validates the effectiveness of our progressive inference approach.

\begin{figure}
  \centering
  \includegraphics[width=4.51cm]{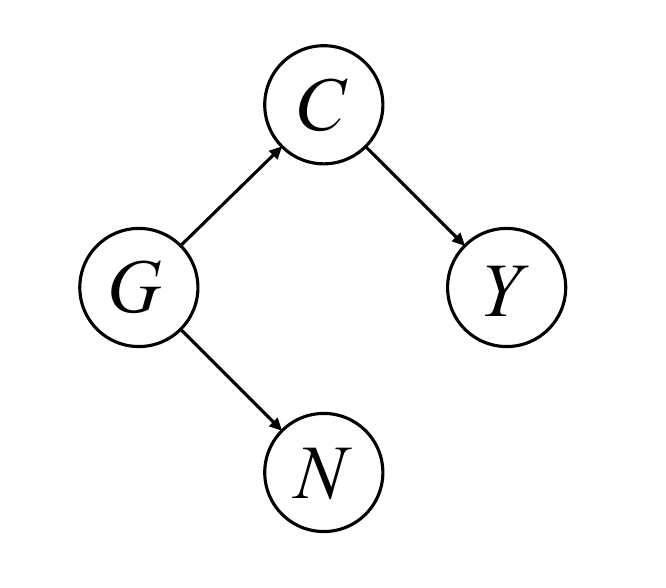}
  \caption{A causal perspective on the graph classification task via structural causal model (SCM).}
  \label{fig:SCM}
\end{figure}

\subsection{GNNs Generalization from A Causal Perspective}
We provide a causal perspective on the graph classification task via a Structural Causal Model (SCM)~\cite{pearl2009causality,peters2017elements}. SCM uses a directed graph to describe the causal relationship between variables, the arrows depict the causal effect. As shown in Figure~\ref{fig:SCM}, we show the causal relationship between 4 variables: the input graph $G$, the ground-truth label $Y$, the causal variable $C$ and the non-causal variable $N$ in the graph. We list a more detailed explanation of the arrows in the SCM:

$\bullet$ $C\leftarrow G \rightarrow N$. The input graph consists of causal variables $C$ and non-causal variables $N$. The causal variable $C$ reflects the intrinsic property characteristics of the graph data, and the non-causal variable $N$ does not determine the intrinsic property. 
For example, in the molecular graph, the chemical properties of organic compounds are mainly determined by their functional groups, i.e., the functional groups here act as causal variables $C$, while the rest of the molecular structures are non-causal variables $N$ that do not determine the chemical properties. 

$\bullet$ $C \rightarrow Y$. The causal variable $C$ is the only one that determines the ground-truth label $Y$.

By observing Figure~\ref{fig:SCM}, it can be found that there is a backdoor path between N and Y, i.e., $N\leftarrow G \rightarrow C \rightarrow Y$. However, when the non-causal variable $N$ and the causal variable $C$ are associated multiple times on the graph $G$, it leads to the establishment of a spurious correlation between the non-causal variable $N$ and the label $Y$. This spurious correlation introduced by the backdoor path can cause the model to fail in the face of OOD. For example, the model may determine the properties of an organic compound by its main chain, because a certain functional group and a certain main chain usually co-occur in the training set. To improve the generalization ability of GNNs, one possible approach is to distinguish between causal variables $C$ and non-causal variables $N$ in the graph $G$, and eventually, encourage causal invariance between causal variables $C$ and labels, and eliminate spurious correlations between non-causal variables $N$ and labels.

\end{document}